\documentclass[10pt,twocolumn,letterpaper]{article}

\usepackage{cvpr}              

\definecolor{cvprblue}{rgb}{0.21,0.49,0.74}
\usepackage[pagebackref,breaklinks,colorlinks,allcolors=cvprblue]{hyperref}

\usepackage[T1]{fontenc}
\usepackage{url}            
\usepackage{booktabs}      
\usepackage{amsfonts}       
\usepackage{nicefrac}       
\usepackage{microtype}      
\usepackage{xcolor}         
\usepackage{natbib}
\usepackage[accsupp]{axessibility} 
\usepackage{graphicx}
\usepackage{color}
\usepackage{siunitx}
\usepackage{colortbl}
\usepackage{multirow}
\usepackage{mathtools}
\usepackage{adjustbox}
\usepackage{algorithm}
\usepackage{algpseudocode}
\usepackage{pythonhighlight}
\usepackage[symbol]{footmisc}
\usepackage{comment}  
\usepackage{amsfonts}       
\usepackage{makecell}
\usepackage{wrapfig}
\usepackage{caption}
\usepackage[capitalise,capitalize]{cleveref}
\usepackage{kotex}
\usepackage{tcolorbox}

\def\paperID{20762} 
\def\confName{CVPR}
\def\confYear{2026}

\title{Visual-RRT: Finding Paths toward Visual-Goals via Differentiable Rendering}

\author{
Sebin Lee$^{1}$\footnotemark[1]\,\,\,\,\,
Jumin Lee$^{1}$\footnotemark[1]\,\,\,\,\,
Taeyeon Kim$^1$\,\,\,\,\,
Youngju Na$^1$\,\,\,\,\,
Woobin Im$^2$\,\,\,\,\,
Sung-Eui Yoon$^1$
\\\\
$^1$KAIST\,\,\,\,\,\,\,\,\,\,\,\,\,\,\,\,\,\,$^2$Samsung Electronics
\vspace{-2mm}
}

\begin{document}

\maketitle
\footnotetext[1]{Both authors contributed equally to this work as co-first authors.}

\begin{abstract}

Rapidly-exploring random trees (RRTs) have been widely adopted for robot motion planning due to their robustness and theoretical guarantees. However, existing RRT-based planners require explicit goal configurations specified as numerical joint angles, while many practical applications provide goal specifications through visual observations such as images or demonstration videos where precise goal configurations are unavailable. 
In this paper, we propose visual-RRT (vRRT), a motion planner that enables visual-goal planning by unifying gradient-based exploitation from differentiable robot rendering with sampling-based exploration from RRTs. 
We further introduce (i) a frontier-based exploration-exploitation strategy that adaptively prioritizes visually promising search regions, and (ii) inertial gradient tree expansion that inherits optimization states across tree branches for momentum-consistent gradient exploitation. Extensive experiments across various robot manipulators including Franka, UR5e, and Fetch demonstrate that vRRT achieves effective visual-goal planning in both simulated and real-world settings, bridging the gap between sampling-based planning and vision-centric robot applications. Our code is available at \url{https://sgvr.kaist.ac.kr/Visual-RRT}.
\vspace{-3mm}

\end{abstract}

\section{Introduction}
\begin{figure*}[]
  \centering
    \includegraphics[width=\linewidth,trim={0cm 7.5cm 0cm 0cm},clip]{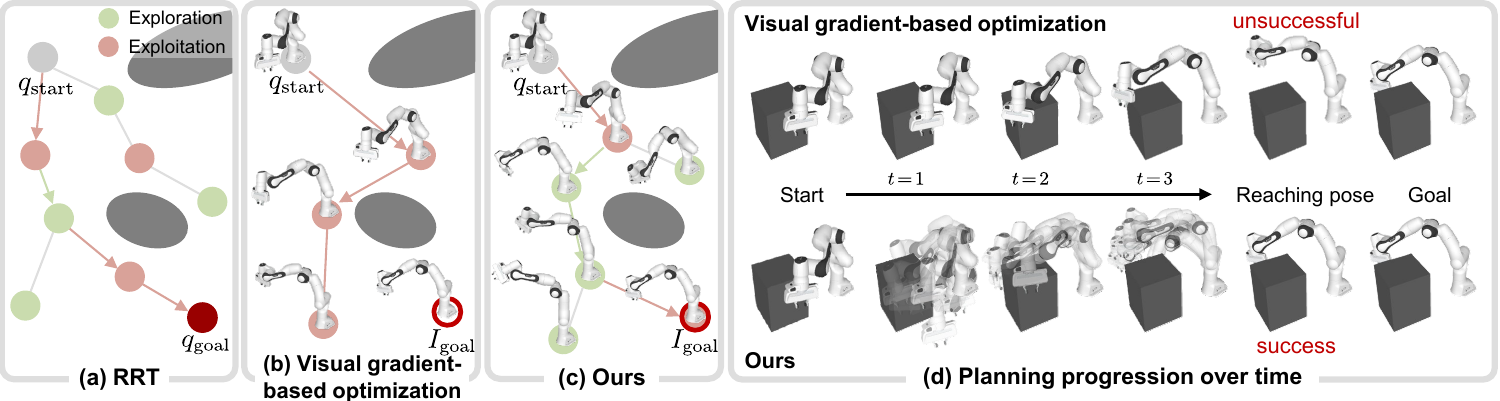}
        \vspace{-7mm}
\caption{\textbf{Motion planning toward visual goals.}
(a) RRT planners efficiently explore the C-space from the start configuration $q_\text{start}$ yet require explicit goal configurations $q_\text{goal}$, limiting their direct use with image-specified goals.
(b) Visual gradient-based methods minimize the rendering loss with respect to the goal image $I_\text{goal}$, but often struggle to reach the desired configuration.
(c) Our vRRT integrates sampling-based exploration and visual-gradient exploitation, enabling the planner to efficiently discover a path toward $I_\text{goal}$.  
Gray regions in (a–c) indicate unreachable parts of the C-space.  
(d) Planning progression over time: while gradient-based optimization stagnates in local minima (top), our method continues to explore and successfully reaches the visual goal (bottom).
In each step of our visualization, multiple robot poses represent parallel exploration and exploitation directions sampled during a single expansion step.
}
\label{fig:intro} \vspace{-4mm}
\end{figure*}

Motion planning is a fundamental capability for autonomous robot manipulators, enabling them to find motion paths from a start to a goal configuration~\cite{prm, stmop, chomp, trjopt}. Among these approaches, the rapidly-exploring random tree (RRT)~\cite{rrt} and its variants~\cite{rrtstar, rrtconnect, infromedrrtstar} have become cornerstone methods in robotics owing to their robustness and theoretical guarantees (\textit{e.g.}, probabilistic completeness or asymptotic optimality~\cite{solovey2020revisiting}). 
To find motion paths, RRTs incrementally expand a search tree toward randomly sampled configurations (Fig.~\ref{fig:intro}a), efficiently exploring the configuration space (C-space) and bypassing local minima.
Building on this, various works have improved path quality~\cite{rrtstar,solovey2020revisiting}, convergence speed~\cite{rrtconnect,quickrrtstar}, and sampling efficiency~\cite{infromedrrtstar,prrtstar,dynamic_domain_rrt} of RRTs.

Despite their advances, RRT-based motion planners assume that explicit goal configurations are given, typically as numerical joint angles. 
In contrast, several emerging applications define robot goals through visual observations such as images or demonstration videos where the desired goal configuration is not explicitly known~\cite{drrobot,profrobot}. 
However, existing RRT variants do not directly accommodate visually defined goals, limiting their applicability in vision-centric robotics where only goal images are provided.

To bridge this gap, we extend RRTs to directly handle visual goals without explicit goal configurations. 
However, this extension is non-trivial since RRTs rely not only on exploration but also on exploitation where explicit goal configurations are crucial to guide goal-directed tree growth (\textit{e.g.}, goal biasing~\cite{rrt}, bi-directional search~\cite{rrtconnect,st_rrtstar,bi_am_rrtstar}, and potential field construction~\cite{prrtstar,apf_rrtstar}).
Differentiable robot rendering~\cite{drrobot,profrobot} has shown that visual gradients of rendering loss enable such exploitation (Fig.~\ref{fig:intro}b), offering a promising foundation for visual-goal planning.

In this paper, we propose visual-RRT (vRRT), a visual-goal motion planner that unifies gradient-based exploitation from differentiable robot rendering~\cite{drrobot} with sampling-based exploration from RRTs~\cite{rrtstar} (Fig.~\ref{fig:intro}c).
While differentiable rendering provides visual gradients, their single-path nature is prone to local minima (Fig.~\ref{fig:intro}d), requiring strategic integration with RRT's multi-branch tree structure for effective goal-directed planning.
Our key idea is to guide tree expansion toward promising configurations using visual gradients while maintaining stochastic sampling for global coverage.
To further improve this combination, we introduce a frontier-based exploration-exploitation strategy that adaptively prioritizes visually promising search regions, leading to more effective planning.
In addition, we propose inertial gradient tree expansion to formulate gradient-based exploitation, as existing RRT states (\textit{e.g.}, path cost) are not designed to capture gradient optimization history. Here, each expansion inherits optimization states from its parent for continuous optimization, enabling momentum-consistent search progress across tree branches while preserving RRT's exploration capability. In experiments, we validate vRRT across various robot platforms with both real and synthetic datasets, demonstrating its effectiveness in visual-goal motion planning. 

\noindent Our contributions are summarized as follows:
\begin{itemize}
    \item We propose vRRT that unifies gradient-based exploitation from differentiable rendering with sampling-based exploration from RRTs, enabling visual-goal motion planning without explicit goal configurations.
    \item We introduce (i) a frontier-based exploration-exploitation strategy that adaptively prioritizes visually promising search regions, along with (ii) inertial gradient tree expansion that inherits optimization states across tree branches for momentum-consistent gradient exploitation.
    \item Extensive experiments demonstrate that vRRT consistently outperforms existing visual-goal planners in both simulated and real-world settings with various robot manipulators including Franka, UR5e, and Fetch.
\end{itemize}
\section{Related Work}
\vspace{0mm}
\noindent \textbf{Rapidly-exploring Random Tree (RRT)}
and its variants~\cite{rrtstar, rrtconnect, infromedrrtstar} have become widely adopted in robotic manipulation due to their robustness and theoretical guarantees (\textit{e.g.}, asymptotic optimality~\cite{solovey2020revisiting}). 
Their core principle is to grow a search tree by sampling configurations while balancing exploration with goal-biased exploitation.
Existing RRTs have enhanced path quality or planning efficiency through tree rewiring~\cite{rrtstar, solovey2020revisiting}, bi-directional tree expansion~\cite{rrtconnect,bi_am_rrtstar,st_rrtstar}, and goal-directed biasing via potential fields~\cite{prrtstar,apf_rrtstar,dynamic_domain_rrt}. 
Recently, learning-guided RRTs have also shown promising planning results by conducting planning in latent spaces~\cite{l2rrt} or by replacing heuristic components with neural samplers~\cite{mpnet_tro,qureshi2018deeply,neuralrrtstar,mpnet,neural_informed_rrtstar}.
However, both classical and learning-guided RRTs often rely on explicit goal configurations for effective tree expansion.

In this paper, we introduce a visual-RRT that redefines the planning objective of RRTs in the visual domain. 
Rather than exploiting known goal configurations, we propose to leverage differentiable robot rendering to guide tree expansion: visual-gradient steering provides exploitation signals by guiding branches toward poses whose renderings align with the goal image. This unification of gradient-based and sampling-based approaches enables effective visual-goal motion planning without explicit goal configurations.

\noindent \textbf{Differentiable Robot Rendering.} $\,$
Recent advances in differentiable rendering~\cite{kerbl2023, wu20234d, Huang2DGS2024, li20253d, zhang2025eggs} have been applied to various robotics domains thanks to their photorealistic 3D scene representations: \textit{e.g.}, world modeling~\cite{lu2024manigaussian, lu2025gwm}, control~\cite{li2025controlling}, and navigation~\cite{chen2025splat, adamkiewicz2022vision, chen2025control}. 
Along with these advances, differentiable robot rendering~\cite{drrobot,profrobot} fully leverages this differentiability to establish visual gradient-based connections between robot configurations and rendered images.
Their core idea is to construct a self-model of the robot's own body by combining Gaussian Splatting~\cite{kerbl2023}, forward kinematics, and implicit linear blend skinning~\cite{loper2023}, enabling rendering losses to be backpropagated to joint angles.
While these methods support vision-centric robot applications (\textit{e.g.}, self-pose reconstruction and text-to-pose generation), they rely on single-path optimization that is prone to local minima.

In this work, we address this challenge by integrating visual gradients from differentiable rendering with sampling-based exploration from RRTs. This fusion enables gradient-based goal-directed progress while maintaining multiple search paths through RRT's tree structure, enabling more robust planning. To realize this integration, we introduce adaptive strategies that balance exploration and exploitation across the tree, enabling momentum-consistent gradient descent within RRT's probabilistic framework. To the best of our knowledge, this is the first method that integrates differentiable rendering-based visual gradients directly into a sampling-based planner such as RRT, extending the scope of differentiable rendering to robot planning scenarios.

\begin{figure*}[t!]
  \centering
    \includegraphics[width=\linewidth,trim={0cm 7.66cm 0cm 0cm},clip]{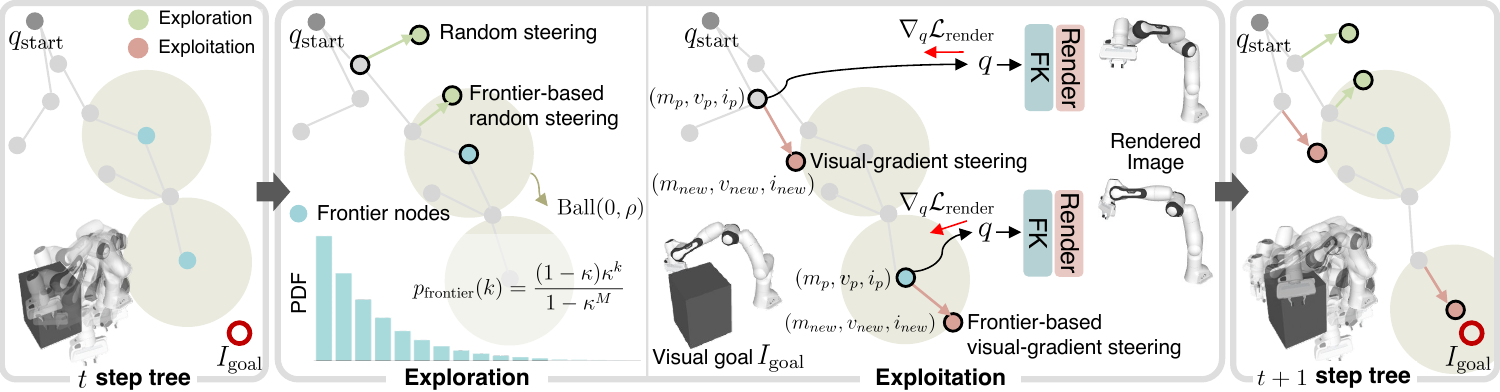}
        \vspace{-6.5mm}
\caption{\textbf{Overview of visual-RRT (vRRT).} vRRT incrementally grows a search tree from step $t$ to $t+1$ toward a visual goal $I_\text{goal}$ by integrating sampling-based exploration with gradient-based exploitation. 
\textbf{Frontier-based steering:} At each iteration, visually promising nodes (cyan) are prioritized as parents via a truncated geometric distribution $p_\text{frontier}(k)$ to balance goal-directed exploitation with exploration.
\textbf{(Exploration)} Random steering expands the tree toward randomly sampled configurations from $\text{Ball}(0, \rho)$ around selected frontier nodes for broad C-space coverage. 
\textbf{(Exploitation)} Visual-gradient steering leverages differentiable rendering: parent configurations $q$ are rendered to compute visual loss $\mathcal{L}_\text{render}$ against $I_\text{goal}$, yielding visual gradient $\nabla_q \mathcal{L}_\text{render}$ for goal-directed updates. 
We introduce inertial gradient expansion where each node stores optimization 
states (first moment $m_p$, second moment $v_p$, iteration $i_p$) inherited 
from its parent, enabling independent momentum-based gradient trajectories 
across multiple branches toward the visual goal.}
\label{fig:method}\vspace{-4mm}
\end{figure*}

\section{Method}

\subsection{Visual-RRT for Visual-Goal Motion Planning}

A robot manipulator's configuration $q$ is defined within its configuration space (C-space) $\mathcal{Q} \subset \mathbb{R}^d$ with $d$ degrees of freedom (\textit{e.g.}, the number of joints).
In the C-space, visual-goal motion planning aims to seek a motion path consisting of configurations $\tau = (q_\text{start}, q_1, \ldots, q_T)$ such that $q_\text{start}$ is a given start configuration, and $q_T$ visually corresponds to the robot pose depicted in a given goal image $I_\text{goal}$. 

As shown in Fig.~\ref{fig:method}, our vRRT incrementally grows a search tree in the C-space by performing exploration and exploitation at each iteration $t$. Starting from the start configuration $q_\text{start}$, we expand the tree in two complementary ways: exploration via random steering maintains global C-space coverage by extending toward randomly sampled configurations, while exploitation via visual-gradient steering guides promising nodes toward the visual goal $I_\text{goal}$ using gradients from differentiable rendering. At each iteration, both strategies operate in parallel on a batch of parent nodes\footnote{Batch indices are omitted for notational simplicity.}, enabling the tree to explore diverse regions while progressively converging to visually matching configurations. 

\vspace{0.3mm}
\noindent \textbf{Random Steering and Visual-Gradient Steering.}
At each planning iteration $t$, we sample a batch of random configurations from the C-space and identify their nearest nodes in the current tree as parent nodes following standard RRT~\cite{rrt}.
We then expand these parents using two strategies: random steering for exploration and visual-gradient steering for exploitation.
For random steering, given a parent $q_p$ and its corresponding random sample $q_\text{rand}$, we steer toward $q_\text{rand}$ with a fixed step size $\epsilon$, generating a child node $q_\text{new} = q_p + \epsilon \cdot \frac{q_\text{rand} - q_p}{\|q_\text{rand} - q_p\|}$.
For visual-gradient steering, we propose leveraging gradient-based configuration optimization of differentiable robot rendering~\cite{drrobot,profrobot}: given a parent $q_p$, we render the robot image $I(q_p) = \pi(\text{FK}(q_p))$, where robot forward kinematics $\text{FK}$ and renderer $\pi$ form a differentiable pipeline from configuration to image space.
We then compute the rendering loss $\mathcal{L}_\text{render}(q_p) \! =  \! \|I(q_p) - I_\text{goal}\|$ to the goal image.
Accordingly, we obtain $\nabla_q \mathcal{L}_\text{render}(q_p)$ and steer toward visually aligned configurations: 
\vspace{-1mm}\begin{equation}
    q_\text{new} = q_p - \alpha \cdot \nabla_q \mathcal{L}_\text{render}(q_p),    
\end{equation}
where $\alpha$ is the gradient step size.
Newly generated nodes are added to the tree if they satisfy collision-free constraints, allowing the tree to grow both broadly (exploration) and directionally (exploitation).

\subsection{Frontier-based Exploration-Exploitation}

While random steering and visual-gradient steering enable tree expansion, naively applying them to all nodes leads to inefficient planning.
Efficient RRT-based planning~\cite{rrtstar} relies on goal biasing to balance exploration and exploitation by prioritizing growth toward known goal configurations. However, this principle poses a challenge for visual-goal planning, where explicit goal configurations are unavailable. 
To address this, we introduce a frontier-based strategy that extends goal biasing to the visual domain.
Our key insight is that visual loss $\mathcal{L}_\text{render}$ 
serves as an implicit goal proximity measure: nodes with lower loss are more likely to approach the visual goal. We therefore adaptively prioritize promising node selection based on visual loss, effectively implementing visual goal biasing while maintaining sufficient exploration.

At each iteration $t$, we maintain a frontier set $\mathcal{F}_t$ consisting of nodes in the current tree, where lower loss indicates higher visual promise.
We rank the top $M$ frontier nodes by their visual loss in ascending order: $\{q_0, q_1, \ldots, q_{M-1}\}$ where $\mathcal{L}_\text{render}(q_i) \leq \mathcal{L}_\text{render}(q_{i+1})$.
Rather than uniformly sampling from all frontier nodes or greedily selecting the best node, we employ a truncated geometric distribution over ranks to balance exploration and exploitation:
\vspace{-1mm}\begin{equation}
    p_\text{frontier}(k) = \frac{(1-\kappa)\kappa^{k}}{1-\kappa^{M}}, \quad k \in \{0, 1, \ldots, M-1\},
\end{equation}
where $\kappa \in [0, 1)$ controls the selection bias toward low-loss nodes.
With this distribution, nodes with lower visual loss (lower rank) have higher selection probability, while still allowing exploration of less promising nodes.

We sample multiple parent nodes from $\mathcal{F}_t$ according to $p_\text{frontier}(k)$, and for each sampled parent, we perform two types of expansion independently: frontier-based \textit{exploration} and gradient-based \textit{exploitation}.
For frontier-based exploration, we sample a local target configuration $q_\text{target} = q_f + u$, where $q_f$ is a node sampled from the frontier according to $p_\text{frontier}(k)$ and $u \sim \text{Uniform}(\text{Ball}(0, \rho))$ is uniformly sampled from a $d$-dimensional ball of radius $\rho$. We then identify the nearest node to $q_\text{target}$ in the tree as the parent $q_p$, and perform random steering toward the local target: $q_\text{new} = q_p + \epsilon \cdot \frac{q_\text{target} - q_p}{\|q_\text{target} - q_p\|}$.
This anchored sampling strategy provides a soft goal bias by exploring the neighborhood of visually promising nodes while maintaining stochastic diversity.
For gradient-based exploitation, we apply visual-gradient steering: $q_\text{new} = q_p - \alpha \cdot \nabla_q \mathcal{L}_\text{render}(q_p)$.
The newly generated nodes are added to the tree, and the frontier set $\mathcal{F}_{t+1}$ is updated accordingly.

Our frontier-based method balances exploration and exploitation through its rank-based sampling: while the distribution's shape controlled by $\kappa$ remains fixed throughout planning, the frontier set $\mathcal{F}_t$ evolves as the tree discovers nodes with lower visual loss.
Consequently, sampling from the same truncated geometric distribution increasingly targets visually promising regions of the C-space as planning progresses, while still allocating probability mass to higher-rank nodes for exploration.
This allows vRRT to effectively shift its computational focus toward goal-relevant configurations without explicitly modifying the sampling strategy.

\subsection{Inertial Gradient Tree Expansion}

While visual-gradient steering enables gradient-based exploitation in tree expansion, effectively leveraging gradient descent within RRT's tree structure requires careful consideration of optimization dynamics.
Unlike traditional gradient descent that follows a single trajectory with accumulated momentum, tree-based planning explores multiple paths simultaneously, where each branch represents a potential solution.
Although some RRT variants maintain states such as path cost~\cite{rrtstar} or kinodynamic constraints~\cite{kynodynamic_rrt, ortiz2024idb}, these states are not designed to capture the optimization history necessary for effective gradient-based convergence.
Consequently, each gradient step in conventional tree expansion resets the optimization, leading to inefficient convergence and sensitivity to local minima.

To address this, we propose inertial gradient tree expansion, where each tree node maintains its own optimization trajectory by inheriting gradient descent states from parent to child.
This allows multiple gradient-based search paths to proceed in parallel across the tree, each with accumulated momentum that guides convergence toward visually promising regions.
Specifically, each node stores optimization states $(m_p, v_p, i_p)$: first moment $m_p$, second moment $v_p$, and iteration step $i_p$, capturing its gradient descent history.
When a child node $q_\text{new}$ is created via visual-gradient steering from parent $q_p$, it inherits and updates these states, enabling continuous optimization along the branch.
We implement the state update using adaptive moment estimation, following the Adam process~\cite{Kingma2015AdamAM}, where the iteration step is incremented as $i_\text{new} \! = \!  i_p \! + \! 1$ and the moments are updated as:
\vspace{-5mm}\begin{align}
    m_\text{new} &= \beta_1 m_p + (1 - \beta_1) \nabla_q \mathcal{L}_\text{render}(q_p), \\
    v_\text{new} &= \beta_2 v_p + (1 - \beta_2) (\nabla_q \mathcal{L}_\text{render}(q_p))^2,
\end{align}
where $\beta_1$ and $\beta_2$ are the exponential decay rates for moment estimates of the visual gradient $\nabla_q \mathcal{L}_\text{render}(q_p)$.
With these updated moments, our visual-gradient steering is defined as:
\vspace{-1mm}\begin{equation}
    q_\text{new} = q_p - \alpha \cdot \frac{\hat{m}_\text{new}}{\sqrt{\hat{v}_\text{new}} + \delta},
\end{equation}
where $\hat{m}_\text{new} = m_\text{new} / (1 - \beta_1^{i_\text{new}})$ and $\hat{v}_\text{new} = v_\text{new} / (1 - \beta_2^{i_\text{new}})$ are bias-corrected moment estimates, and $\delta$ is a small constant for numerical stability.
The child node $q_\text{new}$ inherits the updated states $(m_\text{new}, v_\text{new}, i_\text{new})$, enabling continuous optimization along the branch.

Our state inheritance mechanism allows each branch of the search tree to maintain its own gradient descent trajectory with momentum, enabling descendants of promising nodes to benefit from accumulated optimization history for more efficient convergence.
Importantly, this inertial expansion preserves RRT's exploration capability: different branches maintain independent optimization trajectories, while random steering expansions do not inherit optimization states, ensuring continued exploration of diverse C-space regions.

\section{Experiments}\label{sec:expr}
We evaluate vRRT on two complementary tasks: (i) \emph{visual-goal motion planning}, where the planner generates a collision-free path toward a goal image in cluttered scenes, and (ii) \emph{visual-goal pose reconstruction}, which isolates the difficulty of recovering a final configuration that visually matches the goal. 
Finally, we present \emph{ablation studies} on the robustness of visual guidance under noisy goal hints and the exploration-exploitation balance.
Experiments are conducted on three manipulators: Franka Emika Panda (Franka), UR5e, and Fetch in both simulation and real-world settings.

\noindent \textbf{Implementation Details.}
We implement differentiable robot rendering~\cite{drrobot} using 3D Gaussian 
Splatting~\cite{lu2024manigaussian}, where each robot is represented with 5k--10k 
Gaussian primitives trained on images rendered from MuJoCo~\cite{todorov2012mujoco}. 
Images are rendered at $480 \times 480$ resolution with $L_2$ loss.
vRRT uses random steering step $\epsilon = 0.04$, gradient step $\alpha = 0.04$, geometric parameter $\kappa = 0.9$, exploration radius $\rho = 0.7$, and momentum parameters ($\beta_1 = 0.9$, $\beta_2 = 0.9$).
We expand a batch of 32 nodes per iteration, terminating when the loss change falls below $0.0001$ for 100 consecutive iterations, following Dr.Robot~\cite{drrobot}. 
We adopt tree rewiring from RRT$^\ast$~\cite{rrtstar}, standard path shortcutting~\cite{lavalle2006planning}, and collision checking with MuJoCo~\cite{todorov2012mujoco}.
All experiments run on a single RTX4090.

\subsection{Visual-goal Motion Planning}
\noindent \textbf{Experimental setup.} \label{sec:exp_mp}
We evaluate vRRT on generating collision-free paths toward a goal image $I_g$ in cluttered scenes. For each robot, we construct six environments by randomly placing 5--10 objects. In each scene, given a start configuration $q_s$, we sample reachable goal configurations $q_g$ at varying distances using standard RRT and render the corresponding goal images $I_g$. To assess difficulty, we group tasks into five bins based on the C-space distance $\|q_s - q_g\|_2$: $[0.5, 1.0, \dots, 2.5]$ radians, with 100 tasks per bin.

\noindent \textbf{Baseline methods.}
We compare vRRT with three baselines: Dr.Robot~\cite{drrobot}, Prof.Robot~\cite{profrobot}, and Dr.Robot + RRT$^\ast$. Dr.Robot and Prof.Robot directly optimize the rendering loss in configuration space from $q_s$, while Dr.Robot + RRT$^\ast$ is a two-stage approach that first estimates a goal configuration with Dr.Robot and then plans toward it using RRT$^\ast$.

\noindent \textbf{Evaluation metrics.}
We report three standard motion-planning metrics (Tab.~\ref{tab:mp}): (1) \emph{Success Rate (SR)}, the fraction of trials producing a collision-free path whose final configuration is within 0.05 rad average joint error of $q_g$; (2) \emph{Path Length (PL)}, the cumulative distance traveled in configuration space; and (3) \emph{Planning Time (Time)}, the average computation time for successful trials. PL and Time are computed only over successful trials.
\begin{table}[t]
\renewcommand{\tabcolsep}{0.9mm}
\renewcommand{\arraystretch}{0.75}
\centering
\footnotesize
\caption{\textbf{Visual-goal motion planning results.}
Success rate (SR, \%), planning time (Time, s), and path length (PL, rad) across 
configuration-space distance bins. 
vRRT achieves substantially higher success rates, particularly at larger distances. 
Time and PL are averaged over successful trials; vRRT solves a broader range of problem difficulties compared to baselines.}
\vspace{-3mm}
\label{tab:mp}
\begin{tabular}{lll|ccccc|c}
\toprule
\multicolumn{1}{l}{\multirow{2}{*}{Robot}} &
\multicolumn{1}{l}{\multirow{2}{*}{Method}} &
\multicolumn{1}{l|}{\multirow{2}{*}{Metric}} &
\multicolumn{5}{c|}{Target distance bins (rad)} &
\multicolumn{1}{c}{\multirow{2}{*}{Avg.}} \\
& & \multicolumn{1}{c|}{}&
\multicolumn{1}{c}{$0.5$} &
\multicolumn{1}{c}{$1.0$} &
\multicolumn{1}{c}{$1.5$} &
\multicolumn{1}{c}{$2.0$} &
\multicolumn{1}{c|}{$2.5$} &
\multicolumn{1}{c}{} \\
\midrule
\multirow{12}{*}{Franka}  
    & \multirow{3}{*}{Dr.Robot} & SR      & 59.8 & 21.5 &  9.5 &  4.2 &  1.5 & 19.3 \\
    &                           & Time    &  9.08 &  9.67 &  9.57 & 12.41 &  10.52 & 10.25 \\
    &                           & PL  &  2.75 &  3.59 &  4.26 &  5.32 &  6.06 &  4.40 \\
    \cmidrule{2-9}
    & \multirow{3}{*}{Prof.Robot} & SR     & 71.2 & 27.3 & 12.2 &  5.3 &  2.0 & 23.6 \\
    &                             & Time   & 12.24 & 13.14 & 13.22 & 15.48 & 13.98 & 13.61 \\
    &                             & PL &  1.18 &  1.77 &  2.33 &  3.10 &  3.78 &  2.43 \\
    \cmidrule{2-9}
    & \multirow{3}{*}{\makecell{Dr.Robot \\ + RRT$^\ast$}}  & SR     & 71.5 & 28.3 & 12.2 &  5.2 &  2.0 & 23.8 \\
    &                                                & Time   & 19.13 & 19.67 & 19.07 & 21.22 & 19.31 & 19.68 \\
    &                                                & PL &  0.50 &  0.99 &  1.40 &  1.90 &  2.49 &  1.46 \\

    \cmidrule{2-9}
    & \multirow{3}{*}{Ours} & SR            &   93.0 &   89.0 &   81.7 &   67.7 &   44.7 &   75.2 \\
    &                       & Time          &  17.44 &  19.16 &  21.46 &  23.82 &  26.32 &  21.64 \\
    &                       & PL        &   0.50 &   1.03 &   1.62 &   2.18 &   2.70 &   1.61 \\
\midrule
\multirow{12}{*}{UR5e}  
    & \multirow{3}{*}{Dr.Robot} & SR     & 67.0 & 37.9 & 18.7 &  5.0 &  1.0 & 25.9 \\
    &                           & Time   &  3.56 &  3.45 &  3.61 &  3.94 &  2.32 &  3.38 \\
    &                           & PL &  1.54 &  2.47 &  2.93 &  3.51 &  3.58 &  2.80 \\
    \cmidrule{2-9}
    & \multirow{3}{*}{Prof.Robot} & SR     & 71.0 & 40.2 & 19.6 &  5.3 &  4.0 & 28.0 \\
    &                             & Time   &  6.27 &  6.13 &  6.24 &  6.56 &  5.67 &  6.17 \\
    &                             & PL &  0.71 &  1.28 &  1.83 &  2.28 &  3.01 &  1.82 \\
    \cmidrule{2-9}
    & \multirow{3}{*}{\makecell{Dr.Robot \\ + RRT$^\ast$}}  & SR     & 70.7 & 40.5 & 19.7 &  5.5 &  4.0 & 28.1 \\
    &                                                & Time   & 13.52 & 13.51 & 13.70 & 14.06 & 12.44 & 13.45 \\
    &                                                & PL &  0.50 &  1.02 &  1.52 &  2.00 &  2.55 &  1.52 \\
    \cmidrule{2-9}
    &  \multirow{3}{*}{Ours} & SR            &   87.3 &   87.5 &   85.2 &   76.2 &   62.7 &   79.8 \\
    &                        & Time          &  11.08 &  12.26 &  12.76 &  15.43 &  17.42 &  13.79 \\
    &                        & PL        &   0.50 &   1.02 &   1.58 &   2.14 &   2.65 &   1.58 \\
\midrule
\multirow{12}{*}{Fetch}  
    & \multirow{3}{*}{Dr.Robot} & SR     & 56.7 & 22.7 &  8.7 &  4.3 &  1.0 & 18.7 \\
    &                           & Time   &  4.49 &  5.65 &  6.45 &  5.94 &  6.46 &  5.80 \\
    &                           & PL &  1.47 &  2.19 &  2.89 &  2.76 &  3.43 &  2.55 \\
    \cmidrule{2-9}
    & \multirow{3}{*}{Prof.Robot} & SR     & 63.2 & 28.2 & 12.0 &  5.2 &  1.0 & 21.9 \\
    &                             & Time   &  7.48 &  8.72 &  9.19 & 10.43 &  9.32 &  9.03 \\
    &                             & PL &  0.92 &  1.47 &  2.13 &  2.69 &  2.87 &  2.02 \\
    \cmidrule{2-9}
    & \multirow{3}{*}{\makecell{Dr.Robot \\ + RRT$^\ast$}}  & SR     & 64.5 & 29.2 & 12.3 &  5.2 &  1.0 & 22.4 \\
    &                                                & Time   & 18.81 & 20.33 & 20.98 & 20.17 & 21.67 & 20.39 \\
    &                                                & PL &  0.47 &  0.98 &  1.49 &  1.93 &  2.46 &  1.47 \\
    \cmidrule{2-9}
    &  \multirow{3}{*}{Ours}  & SR            &   90.2 &   85.7 &   81.7 &   64.2 &   45.3 &   73.4 \\
    &                         & Time          &  21.47 &  24.49 &  28.01 &  29.53 &  33.56 &  27.41 \\
    &                         & PL        &   0.53 &   1.12 &   1.68 &   2.22 &   2.82 &   1.68 \\
\bottomrule
\end{tabular}
\vspace{-7mm}
\end{table}

\begin{figure*}[]
  \centering
    \includegraphics[width=0.95\linewidth,trim={0cm 3.3cm 0cm 0cm},clip]{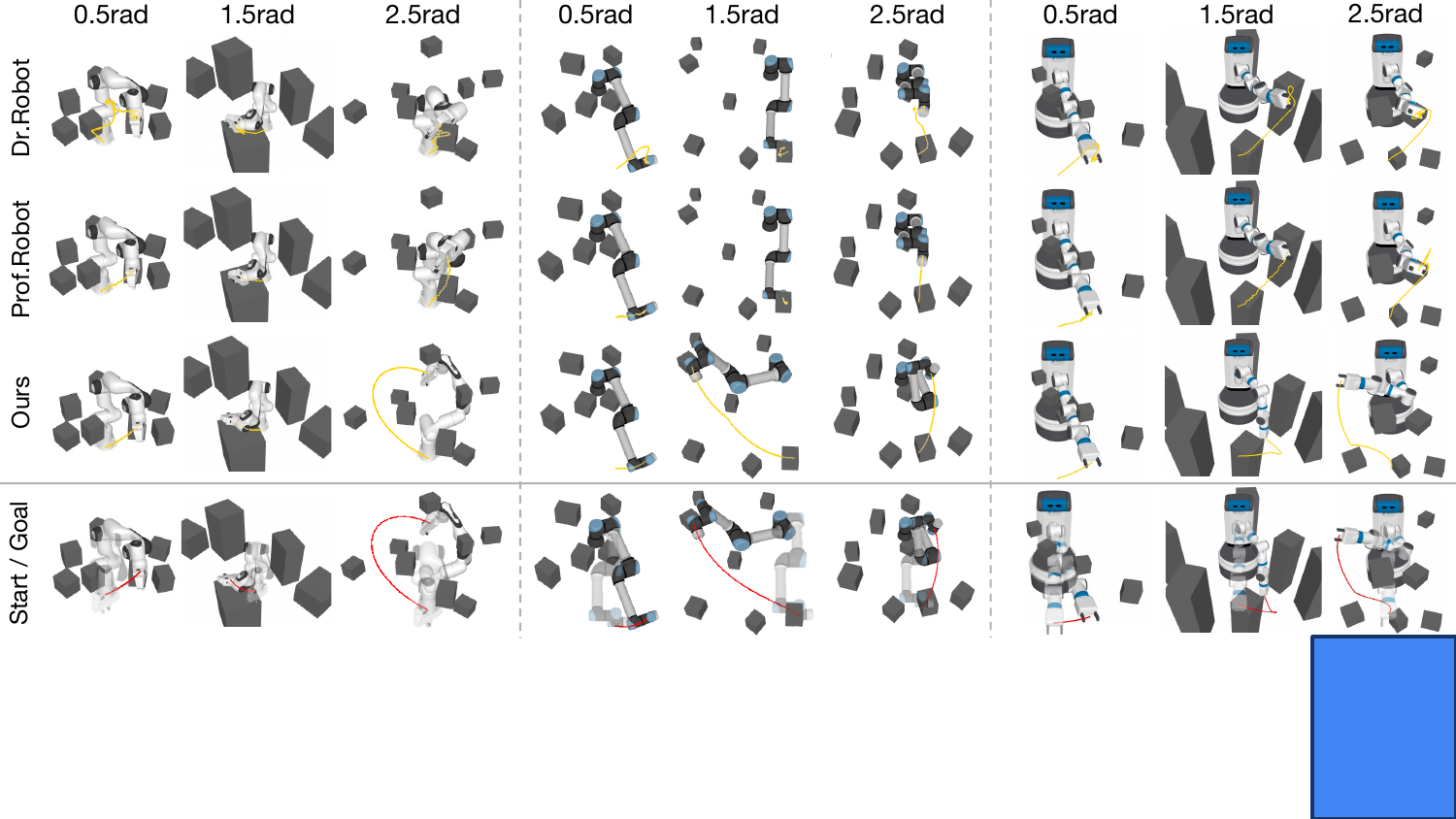}
        \vspace{-2mm}
\caption{\textbf{Qualitative visual-goal motion planning results.}
Top three rows show trajectories (yellow) from Dr.Robot~\cite{drrobot}, Prof.Robot~\cite{profrobot}, and vRRT; 
bottom row shows start/goal poses and RRT$^\ast$ reference (red). 
While all methods succeed at small distances, Dr.Robot produces circuitous trajectories 
and frequently fails at larger distances due to local minima under occlusions. 
Prof.Robot reduces path curvature but remains less effective than vRRT, which consistently 
discovers collision-free paths with geometric structure similar to RRT$^\ast$ solutions.}
    \label{fig:mp}
        \vspace{-5mm}
\end{figure*}

\begin{figure}[]
  \centering
    \includegraphics[width=\linewidth,trim={0cm 0cm 2.1cm 0cm},clip]{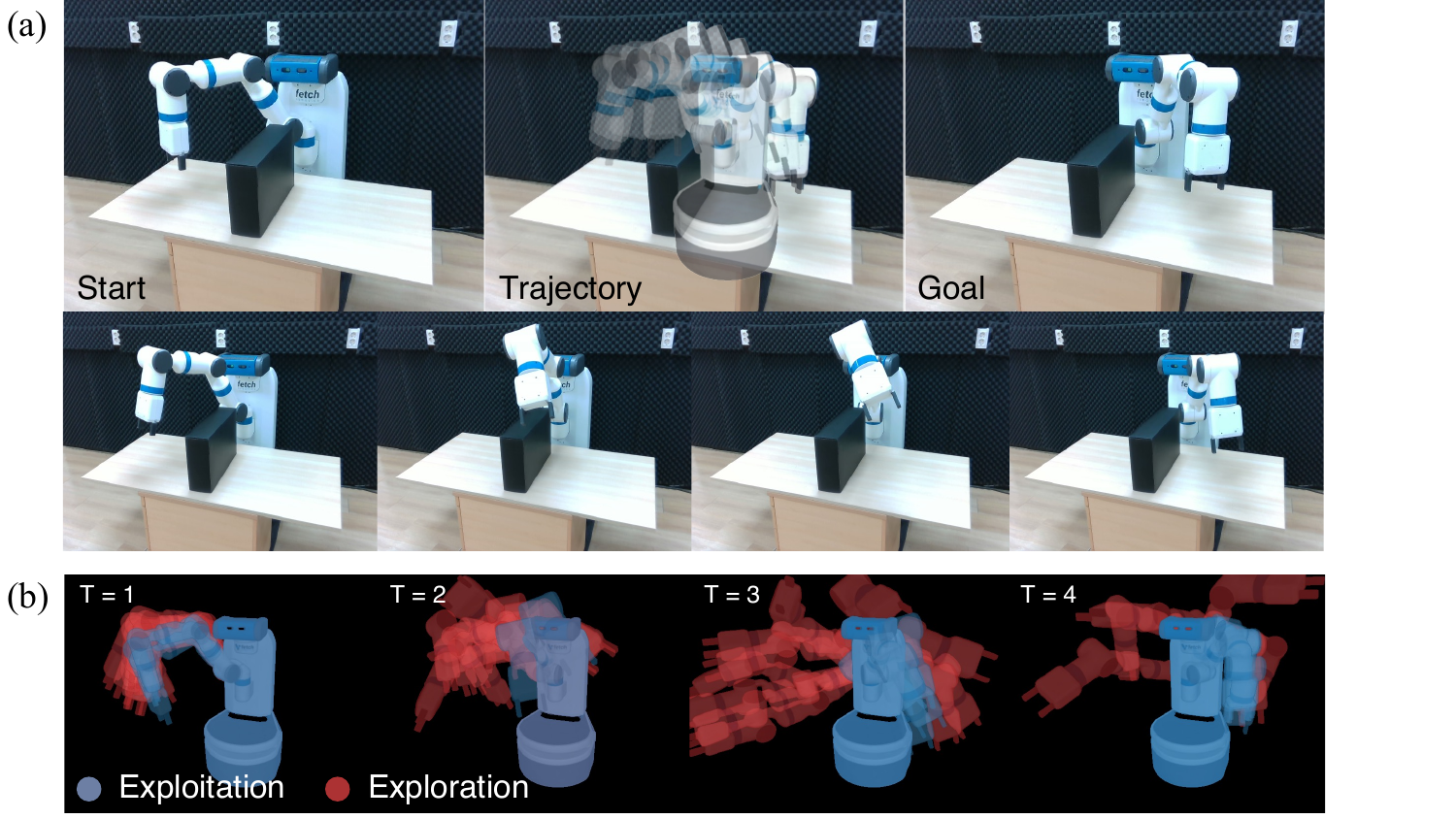}
        \vspace{-7mm}
\caption{\textbf{Real-world deployment on Fetch robot.}
\textbf{(a)} Physical robot execution: given start and goal images (top row), vRRT plans 
a collision-free trajectory executed on the real robot (bottom row frames). 
\textbf{(b)} Planning dynamics: exploitation samples (blue) follow visual gradients while 
exploration samples (red) maintain coverage, demonstrating effective sim-to-real transfer.}
    \label{fig:realfetch}
    \vspace{-6mm}
\end{figure}

\noindent \textbf{Results.}
Tab.~\ref{tab:mp} summarizes motion-planning performance across all robot platforms and difficulty bins. 
The single-path optimization baselines, Dr.Robot and Prof.Robot, degrade substantially as the C-space distance increases. 
In contrast, vRRT achieves the highest success rates on Franka, UR5e, and Fetch, with its advantage becoming more pronounced in harder settings. 
Although the two-stage Dr.Robot + RRT$^\ast$ baseline improves over Dr.Robot alone, it still falls substantially short of vRRT.
This gap arises from error propagation: as discussed in Sec.~\ref{sec:pose_recon}, inaccurate goal estimation in the first stage causes the subsequent RRT$^\ast$ planner to optimize toward a suboptimal target.

Fig.~\ref{fig:mp} further highlights clear differences in path quality. 
Dr.Robot and Prof.Robot frequently fail at larger distances because self-occlusions and scene occlusions induce local minima in gradient-based optimization. 
Even when successful, Dr.Robot tends to generate indirect and circuitous trajectories, while Prof.Robot partially alleviates this issue through smoothing but still produces less direct paths. By contrast, vRRT consistently finds trajectories that closely resemble those of RRT$^\ast$ in geometric structure, indicating that exploration-based tree search can recover efficient motions while remaining guided solely by visual objectives.

In terms of efficiency, vRRT yields moderately longer average path lengths than Dr.Robot + RRT$^\ast$, likely because it succeeds on a broader set of difficult problems.
Planning time remains comparable between the two methods, whereas the single-path optimization baselines run faster but at the cost of substantially lower success rates.

\noindent \textbf{Real-world validation.}
To validate practical applicability and sim-to-real transfer, we deploy vRRT on a physical Fetch robot navigating around tabletop obstacle. 
Fig.~\ref{fig:realfetch}(a) shows a representative trial: given start and goal images, vRRT generates a collision-free path with intermediate poses overlaid in image space, and the robot successfully executes this plan to reach the target configuration. 
Fig.~\ref{fig:realfetch}(b) shows planning dynamics across iterations $T$: exploitation (blue) follows visual gradients while exploration (red) maintains coverage, proving effective under real-world conditions.

\subsection{Visual-goal Pose Reconstruction} \label{sec:pose_recon}
\noindent \textbf{Experimental setup.}
The pose reconstruction task isolates the problem of recovering a configuration that matches a goal image. Each instance consists of a start configuration $q_s$ and a goal image $I_g$ rendered from a target configuration $q_g$. We sample goal configurations across five distance bins, $\|q_s - q_g\|_2 \in [0.5, 1.0, 1.5, 2.0, 2.5]$ rad, with 100 tasks per bin for each robot. We also evaluate on the real-world Panda-3CAM-Azure dataset~\cite{dream}, which provides RGB images of a physical Franka arm with ground-truth joint annotations.

\noindent \textbf{Baselines and metrics.}
We evaluate vRRT against the baselines using three metrics (Tab.~\ref{tab:recon}): (1) \emph{Success Rate (SR)}, the fraction of trials with joint error below 0.05 rad; (2) \emph{Mean Joint Error}, the average per-joint angular deviation; and (3) \emph{PSNR}~\cite{hore2010image}, which measures image-level visual fidelity.
On Panda-3CAM-Azure, we compare vRRT with Dr.Robot~\cite{drrobot} using its warm-start and evaluation protocol, as well as the pose regressors RoboPEPP~\cite{robopepp} and HoRoPose~\cite{horopose}.

\begin{table}[t]
\renewcommand{\tabcolsep}{1.0mm}
\renewcommand{\arraystretch}{0.8}
\centering
\footnotesize
\caption{\textbf{Visual-goal pose reconstruction results.} 
Success rate (SR, \%), mean joint error (Error, rad), and PSNR across configuration-space 
distance bins. 
vRRT demonstrates substantially better performance across all bins in both success rate 
and joint accuracy. 
Notably, vRRT maintains high PSNR--measuring visual similarity, the primary objective in 
visual-goal planning--across distances where Dr.Robot~\cite{drrobot} shows degradation.}
\vspace{-3mm}
\label{tab:recon}
\begin{tabular}{lll|ccccc|c}
\toprule
\multicolumn{1}{l}{\multirow{2}{*}{Robot}} &
\multicolumn{1}{l}{\multirow{2}{*}{Method}} &
\multicolumn{1}{l|}{\multirow{2}{*}{Metric}} &
\multicolumn{5}{c|}{Target distance bins (rad)} &
\multicolumn{1}{c}{\multirow{2}{*}{Avg.}} \\
& & \multicolumn{1}{c|}{}&
\multicolumn{1}{c}{$0.5$} &
\multicolumn{1}{c}{$1.0$} &
\multicolumn{1}{c}{$1.5$} &
\multicolumn{1}{c}{$2.0$} &
\multicolumn{1}{c|}{$2.5$} &
\multicolumn{1}{c}{} \\
\midrule
\multirow{6}{*}{Franka}  
& \multirow{3}{*}{Dr.Robot} & SR    & 92.0  & 53.0  & 21.0  & 11.0  & 2.0   & 35.8 \\
&                           & Error & 0.048 & 0.315 & 0.603 & 0.878 & 1.097 & 0.588 \\
&                           & PSNR  & 30.19 & 26.83 & 23.44 & 22.34 & 21.29 & 24.82 \\
\cmidrule{2-9}
&  \multirow{3}{*}{Ours}    & SR    &  100.0 &   99.0 &   84.0 &   70.0 &   41.0 &   78.8 \\
&                           & Error &  0.015 &  0.019 &  0.098 &  0.159 &  0.420 &  0.142 \\
&                           & PSNR  &  30.84 &  30.92 &  30.10 &  29.39 &  27.82 &  29.81 \\
\midrule
\multirow{6}{*}{UR5e}  
& \multirow{3}{*}{Dr.Robot} & SR    & 80.0  & 57.0  & 32.0  & 19.0  & 7.0   & 39.0 \\
&                           & Error & 0.146 & 0.316 & 0.568 & 0.762 & 1.073 & 0.573 \\
&                           & PSNR  & 32.95 & 30.13 & 28.04 & 26.50 & 24.74 & 28.47 \\
\cmidrule{2-9}
& \multirow{3}{*}{Ours}     & SR    &  100.0 &   98.0 &   92.0 &   74.0 &   55.0 &   83.8 \\
&                           & Error &  0.016 &  0.023 &  0.037 &  0.129 &  0.297 &  0.101 \\
&                           & PSNR  &  34.69 &  34.73 &  34.48 &  34.08 &  33.44 &  34.28 \\
\midrule
\multirow{6}{*}{Fetch}  
& \multirow{3}{*}{Dr.Robot} & SR    &  84.0  & 42.0  & 23.0  & 20.0  & 7.0   & 35.2 \\
&                           & Error & 0.085 & 0.484 & 0.703 & 0.772 & 1.032 & 0.615 \\
&                           & PSNR  & 28.49  & 24.59 & 22.99 & 22.97 & 21.92 & 24.19 \\
\cmidrule{2-9}
&  \multirow{3}{*}{Ours}    &SR    &  100.0 &   91.0 &   76.0 &   59.0 &   35.0 &   72.2 \\
&                           &Error &  0.011 &  0.086 &  0.205 &  0.344 &  0.615 &  0.252 \\
&                           &PSNR  &  29.87 &  29.43 &  28.78 &  28.54 &  27.99 &  28.92 \\
\bottomrule
\end{tabular}
\vspace{-2mm}
\end{table}

\begin{figure}[]
\centering
\includegraphics[width=0.95\linewidth,trim={0cm 0cm 2.6cm 0cm},clip]{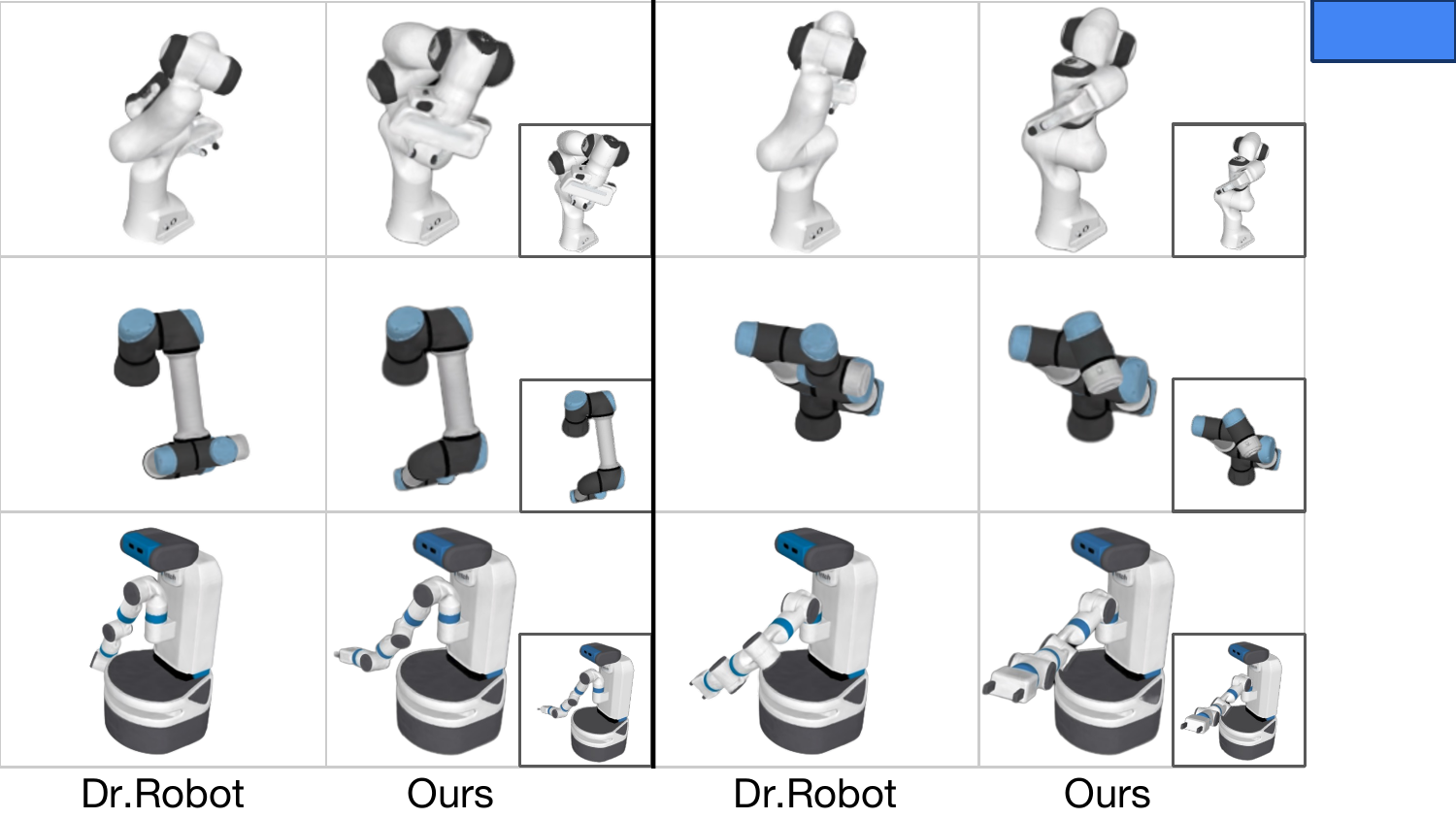}
\vspace{-3mm}
\caption{\textbf{Qualitative pose reconstruction results.}
Reconstruction comparison under self-occlusion (ground truth in insets). 
Under occlusion, Dr.Robot~\cite{drrobot} produces configurations with plausible overall 
silhouettes but incorrect occluded joint angles, whereas vRRT recovers poses more 
consistent with complete robot structure through exploration-based disambiguation.}
\label{fig:recon}
\vspace{-6mm}
\end{figure}

\noindent \textbf{Results.}
Tab.~\ref{tab:recon} summarizes pose reconstruction performance across the three robot platforms. Dr.Robot performs well at small configuration-space distances, demonstrating the effectiveness of gradient-based visual optimization in local regimes. However, its performance degrades steadily as the distance increases. By contrast, vRRT remains robust across all distance bins, achieving higher success rates, lower joint errors, and consistently stronger visual fidelity.

Although joint errors increase with distance for all methods, vRRT maintains high PSNR, indicating that it more reliably recovers configurations that are visually consistent with the goal image. This suggests that tree-based exploration mitigates local minima in direct gradient-based optimization, especially when the target is far in configuration space. Fig.~\ref{fig:recon} supports this observation: Dr.Robot often matches the overall silhouette but fails to recover occluded joint angles, whereas vRRT reconstructs poses that are more consistent with the full robot geometry.

\vspace{0.5mm}
\noindent \textbf{Real-world benchmark evaluation.}
\begin{table}[]
\renewcommand{\tabcolsep}{0.9mm}
\renewcommand{\arraystretch}{0.8}
\centering
\footnotesize
\caption{\textbf{Per-joint reconstruction errors on Panda-3CAM-Azure dataset.}
Comparison of per-joint angular errors (rad) across feed-forward regression methods (RoboPEPP, HoRoPose), gradient-based optimization (Dr.Robot), and vRRT. }
\label{tab:recon-real}
\vspace{-3mm}
\begin{tabular}{lcccccccc}
\toprule
Method & J1 & J2 & J3 & J4 & J5 & J6 & J7 & Avg. \\
\midrule
RoboPEPP~\cite{robopepp} & 0.052 & 0.072 & \textbf{0.068} & 0.206 & 0.125 & 0.579 & --    & 0.184 \\
HoRoPose~\cite{horopose} & 0.193 & 0.123 & 0.197 & 0.054 & \textbf{0.085} & \textbf{0.071} & 0.355 & 0.154 \\
Dr.Robot~\cite{drrobot} & 0.077 & 0.040 & 0.106 & 0.064 & 0.104 & 0.137 & 0.617 & 0.164 \\
Ours   &  \textbf{0.088} & \textbf{0.030} & 0.095 & \textbf{0.030} & 0.162 & 0.082 & \textbf{0.094} & \textbf{0.083} \\
\bottomrule
\end{tabular}
\vspace{-5mm}
\end{table}

\begin{table}[t]
\renewcommand{\tabcolsep}{2.97mm}
\renewcommand{\arraystretch}{0.7}
\centering
\caption{\textbf{Robustness to noisy goal configurations.}
Motion planning success rates (\%) when vRRT uses noisy goal estimates alongside visual objectives--
simulating scenarios with degraded sensors or low-precision encoders. 
High success rates across noise levels demonstrate effective visual-configuration fusion.}
\footnotesize
\vspace{-3mm}
\label{tab:analysis_noisy_q}
\begin{tabular}{c|ccccc|c}
\toprule
\multicolumn{1}{l|}{\multirow{2}{*}{Noise Std.}} &
\multicolumn{5}{c|}{Target distance bins (rad)} &
\multicolumn{1}{c}{\multirow{2}{*}{Avg.}} \\
\multicolumn{1}{c|}{}&
\multicolumn{1}{c}{$0.5$} &
\multicolumn{1}{c}{$1.0$} &
\multicolumn{1}{c}{$1.5$} &
\multicolumn{1}{c}{$2.0$} &
\multicolumn{1}{c|}{$2.5$} &
\multicolumn{1}{c}{} \\
\midrule
 0.05   &   90.7 &   92.8 &   91.0 &   93.0 &   93.8 &   92.3 \\
 0.10   &   89.2 &   91.0 &   91.2 &   91.3 &   90.2 &   90.6 \\
 0.15   &   86.8 &   90.0 &   91.0 &   89.7 &   90.0 &   89.5 \\
 0.20   &   87.8 &   89.5 &   90.5 &   88.3 &   89.0 &   89.0 \\
\bottomrule
\end{tabular}
\vspace{-3mm}
\end{table}
\begin{figure}[]
  \centering
    \includegraphics[width=\linewidth,trim={0cm 5.2cm 0cm 0cm},clip]{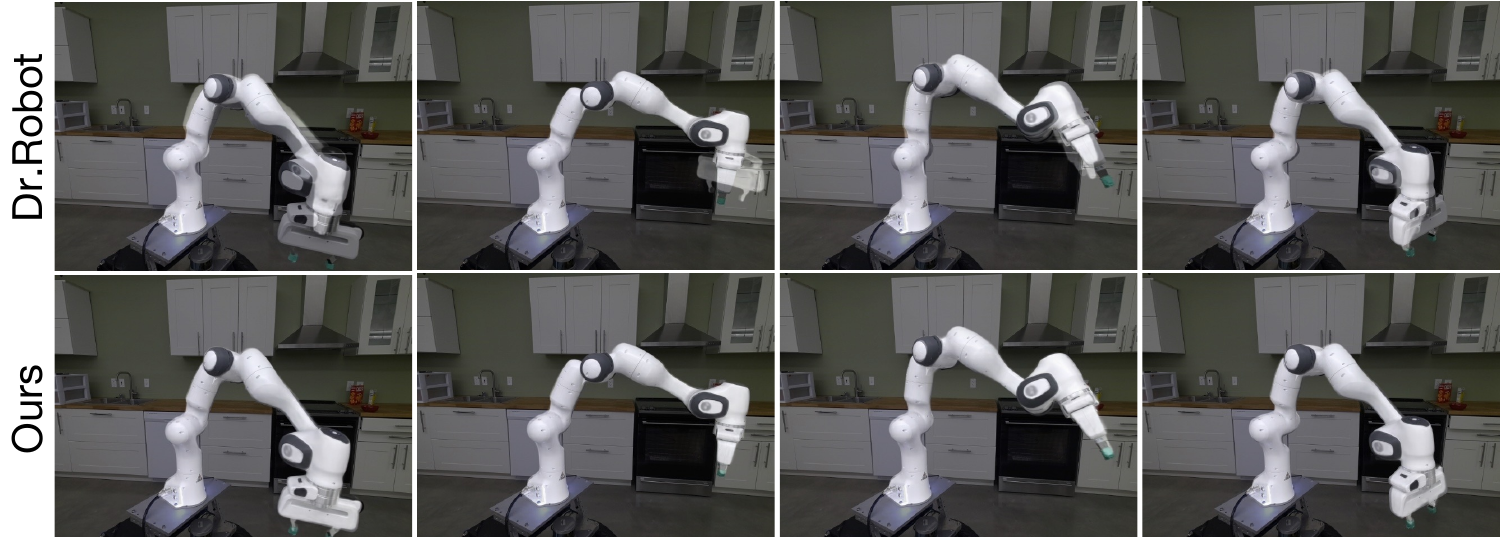}
    \vspace{-7mm}
\caption{\textbf{Qualitative results on Panda-3CAM-Azure dataset.} 
Rendered poses overlaid on goal images demonstrate improved reconstruction quality. 
vRRT achieves better accuracy on occluded joints compared to Dr.Robot~\cite{drrobot}}
    \label{fig:recon-real}
        \vspace{-6mm}
\end{figure}

We evaluate on the Panda-3CAM-Azure benchmark~\cite{dream} to assess generalization to 
real-world images.
We compare against feed-forward pose regressors—RoboPEPP~\cite{robopepp} and 
HoRoPose~\cite{horopose}—fine-tuned on real data.
Tab.~\ref{tab:recon-real} and Fig.~\ref{fig:recon-real} show that, despite direct sim-to-real transfer, 
vRRT achieves the lowest mean error, particularly on occluded J7.
Qualitatively, while Dr.Robot matches overall silhouettes but fails on occluded 
joints, vRRT's exploration resolves these ambiguities.

\subsection{Ablation Study} \label{sec:ablation}
We conduct ablation studies on vRRT's components using the UR5e robot 
with the motion planning setup in Sec.~\ref{sec:exp_mp}.

\vspace{0.5mm}
\noindent \textbf{Robustness to noisy goal configurations.}
Although vRRT is primarily guided by visual objectives, it can also exploit approximate configuration-space goal estimates when available, for example, from low-precision motor encoders, aged sensors, or coarse pose regressors. 
To evaluate robustness to such noisy goal hints, we augment exploitation with configuration-space goal biasing: a subset of parent nodes steers toward a noisy goal configuration $\tilde{q}_g = q_g + \mathcal{N}(0, \sigma^2 I)$ in addition to visual gradient steering. 
Classical RRT methods~\cite{rrt,rrtconnect,rrtstar,l2rrt} would plan solely toward $\tilde{q}_g$, which can lead to incorrect targets under large noise. 
In contrast, vRRT combines noisy configuration hints with visual objectives, allowing visual feedback to compensate for configuration-space errors. 
As shown in Tab.~\ref{tab:analysis_noisy_q}, vRRT maintains high success rates even under substantial noise, demonstrating robustness to imprecise goal estimates.

\begin{figure}[t!]
    \centering
    \includegraphics[width=\linewidth,trim={0cm 3.2cm 0cm 0cm},clip]{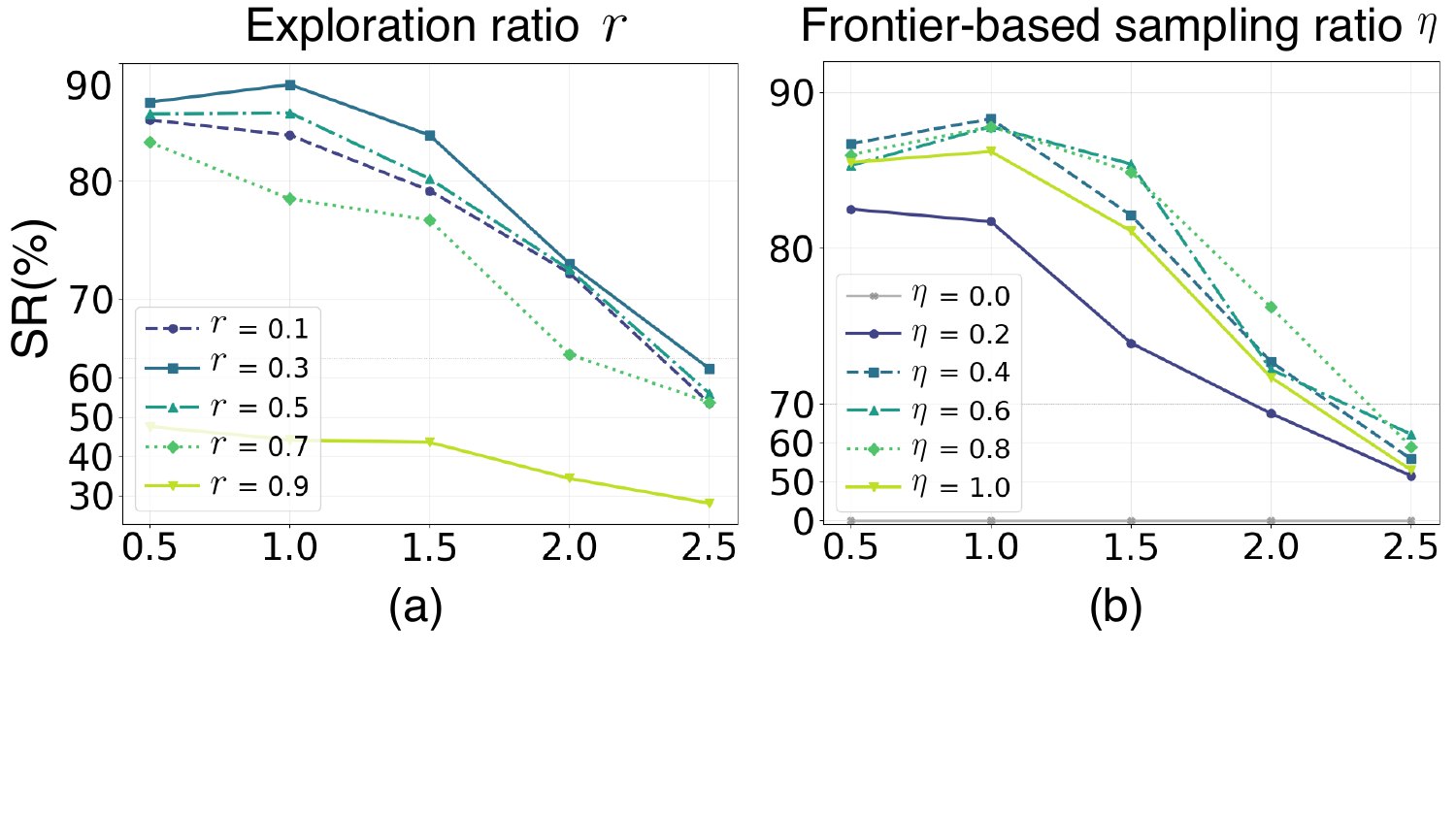}
    \vspace{-8mm}
    \caption{\textbf{Ablation on vRRT tree expansion.}
    Success rates (\%) across distance bins under different (a) exploration ratios $r$ and (b) frontier sampling ratios $\eta$ on UR5e for visual-goal motion planning. 
    The best results are obtained at $r=0.3$ and $\eta=0.6$--$0.8$, indicating that planning requires a balance between goal-directed expansion and exploration. 
    Uniform sampling ($\eta=0.0$) and excessive exploration ($r=0.9$) both degrade performance, confirming the importance of visual guidance and adaptive prioritization.}
    \label{fig:ablation_expl}
    \vspace{-3mm}
\end{figure}
\begin{figure}[]
  \centering
    \includegraphics[width=0.95\linewidth,trim={0.05cm 0.4cm 1.3cm 0cm},clip]{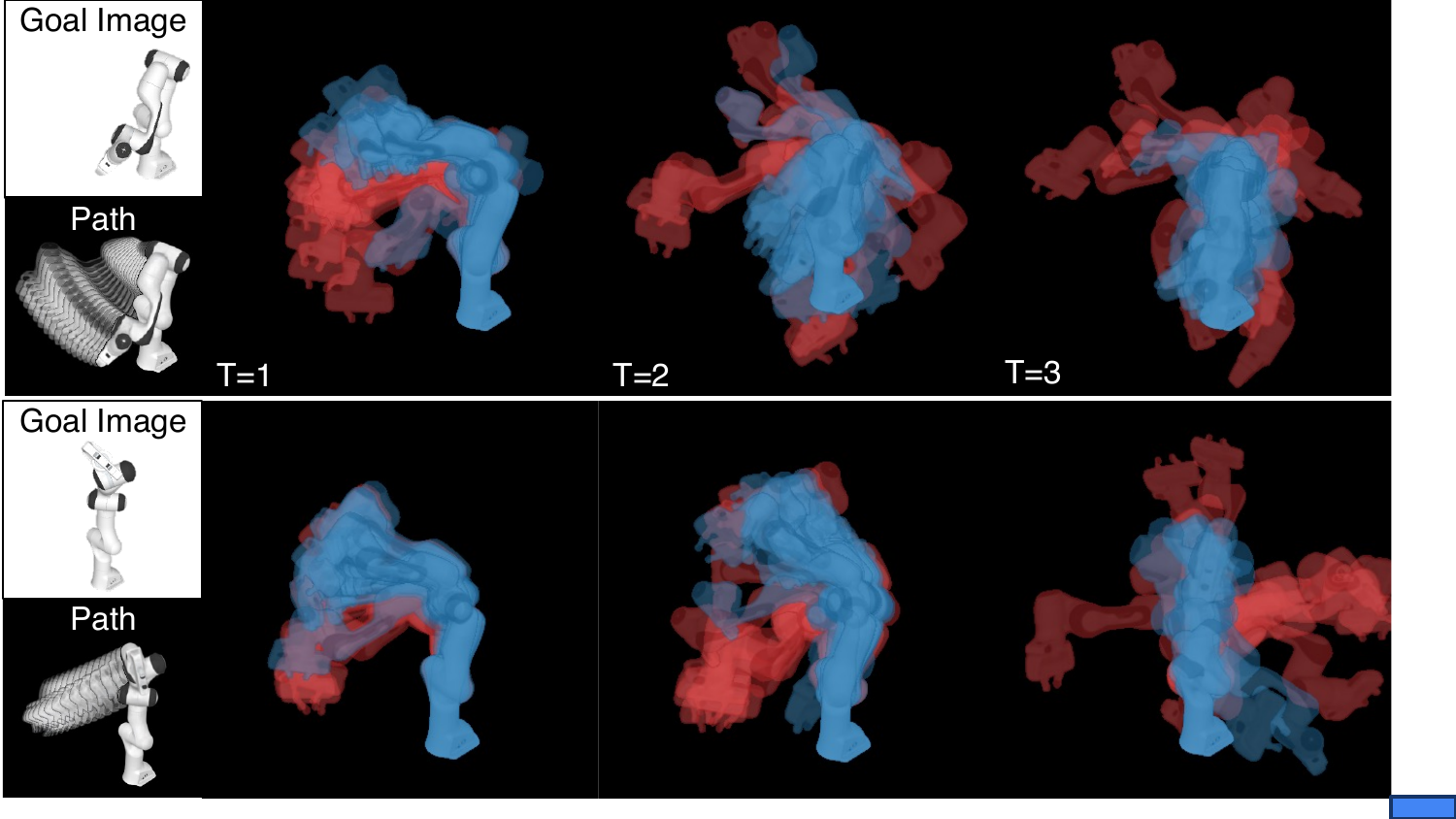}
    \vspace{-3mm}
\caption{\textbf{Planning progression visualization.}
Planning progression at iterations T=1, 2, 3 for two scenarios. 
Blue: exploitation samples following visual gradients. 
Red: exploration samples for coverage. 
The coexistence of both strategies enables vRRT to discover feasible paths while avoiding 
local minima.}\label{fig:qual}
        \vspace{-6mm}
\end{figure}

\vspace{0.5mm}
\noindent \textbf{Exploration-exploitation balance.}
We analyze the effect of exploration ratio $r$, which controls the proportion of parent nodes performing exploration versus exploitation at each iteration. 
Fig.~\ref{fig:ablation_expl}(a) presents success rates across varying $r \in [0.1, 0.3, 0.5, 0.7, 0.9]$.
Excessive exploration ($r=0.9$) severely degrades performance across all distance bins, as random sampling fails to effectively leverage visual guidance toward the goal. 
Conversely, minimal exploration ($r=0.1$) performs reasonably at small distances but exhibits notable degradation at larger distances, where local minima become more prevalent and broader search coverage is beneficial.
Optimal performance is achieved at $r=0.3$, which maintains consistently high success rates across all distances. As shown in Fig.~\ref{fig:ablation_expl}(a), performance gradually degrades as $r$ increases beyond 0.5, confirming that while stochastic exploration prevents local minima entrapment, visual gradient guidance remains critical for effective convergence. 
This exploration-exploitation mechanism is visualized in Fig.~\ref{fig:qual}, where blue and red nodes represent gradient-guided and random steering, respectively.

\vspace{0.5mm}
\noindent \textbf{Frontier-based tree expansion.}
We analyze the frontier sampling ratio $\eta$, which determines the proportion of parent nodes sampled from the frontier set $\mathcal{F}_t$ versus uniform sampling. 
As shown in Fig.~\ref{fig:ablation_expl}(b), without frontier-based sampling ($\eta=0.0$), the planner fails to converge, achieving near-zero success rates. This demonstrates that uniform node selection disperses computational effort unproductively without prioritizing 
visually promising regions.
Even minimal frontier sampling ($\eta=0.2$) recovers performance to 50–80\%, with further improvement at moderate concentrations ($\eta=0.6$, $0.8$). 
Interestingly, exclusive frontier sampling ($\eta=1.0$) shows slight degradation, suggesting that retaining some uniform sampling maintains beneficial diversity. 
These results confirm that adaptive node prioritization is essential while preserving sufficient exploration diversity.
\begin{table}[t]
\renewcommand{\tabcolsep}{2.62mm}
\renewcommand{\arraystretch}{0.7}
\centering
\footnotesize
\caption{\textbf{Effect of Adam moments parameters.}
Success rates (\%) with varying $\beta_1$ and $\beta_2$ in inertial gradient expansion, evaluated on the UR5e robot under the visual-goal motion planning. 
Optimal performance at standard Adam values ($\beta_1=0.9$, $\beta_2=0.9$).}
\vspace{-3mm}
\label{tab:adam}
\begin{tabular}{cc|ccccc|c}
\toprule
\multicolumn{1}{c}{\multirow{2}{*}{$\beta_1$}} &
\multicolumn{1}{c|}{\multirow{2}{*}{$\beta_2$}} &
\multicolumn{5}{c|}{Target distance bins (rad)} &
\multicolumn{1}{c}{\multirow{2}{*}{Avg.}} \\
&\multicolumn{1}{c|}{}&
\multicolumn{1}{c}{$0.5$} &
\multicolumn{1}{c}{$1.0$} &
\multicolumn{1}{c}{$1.5$} &
\multicolumn{1}{c}{$2.0$} &
\multicolumn{1}{c|}{$2.5$} &
\multicolumn{1}{c}{} \\
\midrule
 0.5  &  0.9    &   38.7 &   29.4 &   26.9 &   25.7 &   27.5 &   29.6 \\
 0.7  &  0.9    &   76.5 &   68.0 &   66.3 &   59.0 &   49.0 &   63.8 \\
 0.9  &  0.9    &   87.3 &   87.5 &   85.2 &   76.2 &   62.7 &   79.8 \\
 0.99 &  0.9   &   78.0 &   70.7 &   67.6 &   61.7 &   47.8 &   65.2 \\
 0.9  &  0.99    &   87.0 &   88.3 &   83.7 &   77.5 &   58.0 &   78.9 \\
 0.9  &  0.999    &   86.8 &   88.3 &   85.6 &   73.8 &   61.0 &   79.1 \\
\bottomrule
\end{tabular}
\vspace{-6mm}
\end{table}

\setlength{\columnsep}{5pt} 
\begin{wrapfigure}{r}{0.25\linewidth}
  \centering
    \vspace{-4mm}
  \includegraphics[trim={0cm 5.12cm 21.62cm 0cm},clip,width=\linewidth]{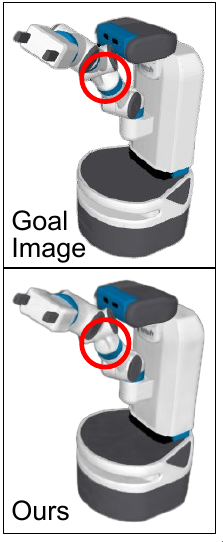}
  \vspace{-7mm}
  \captionof{figure}{\\ \textbf{Limitation.}}
  \label{fig:limitation}
    \vspace{-1mm}
\end{wrapfigure}
\vspace{0.5mm}
\noindent \textbf{Momentum parameter sensitivity.}
We analyze the sensitivity to the optimization moments parameters ($\beta_1$, $\beta_2$) when optimization states are inherited across tree branches.
Tab.~\ref{tab:adam} shows that performance critically depends on $\beta_1$: low momentum ($\beta_1=0.5$) achieves only 29.6\%, while standard momentum ($\beta_1=0.9$) reaches 79.8\%. This substantial gap demonstrates the importance of momentum accumulation for effective gradient exploitation. Excessive momentum ($\beta_1=0.99$) degrades to 65.2\%, likely from over-smoothing gradients. In contrast, $\beta_2$ exhibits robustness across values.

\section{Conclusion}
\vspace{-1mm}
We propose visual-RRT (vRRT), which extends RRT-based motion planning to directly handle visual goals without explicit goal configurations. By unifying gradient-based exploitation from differentiable robot rendering with sampling-based exploration, vRRT enables effective visual-goal planning where only goal images are provided. Our frontier-based exploration-exploitation strategy and inertial gradient tree expansion guide tree growth toward visually promising regions while maintaining diverse search coverage across C-space. Extensive experiments in both simulated and real-world settings validate the effectiveness of vRRT for visual motion planning. We bridge sampling-based planning and vision-centric robotics, offering a direction for integrating differentiable rendering with RRT-based planners.

\noindent \textbf{Limitation.} While vRRT's exploration effectively handles occlusions, visual ambiguity remains challenging. In Fig.~\ref{fig:limitation}, occluded or symmetric robot parts can produce visually similar renderings despite different joint configurations.

\noindent \textbf{Acknowledgement.}
This work was~supported~by~\mbox{Institute} of Information \& Communications~Technology~\mbox{Planning} \& \mbox{Evaluation (IITP) and the National Research Foundation} (NRF), \mbox{funded by the Korea government (MSIT): RS-2023-} 00237965 and RS-2023-00208506.

\clearpage
\appendix
\setcounter{table}{0}
\setcounter{figure}{0}
\setcounter{equation}{0}
\makeatletter
\renewcommand{\thefigure}{S\@arabic\c@figure}
\renewcommand{\thetable}{S\@arabic\c@table}
\renewcommand{\theequation}{S\@arabic\c@equation}
\makeatother

\section*{\Large Supplementary Material} 
\addcontentsline{toc}{section}{Supplementary Material}

In this supplementary material, we provide additional technical details and 
experimental results to complement the main paper. Sec.~\ref{suppl:rrt} 
describes the background of RRT planners. Sec.~\ref{suppl:ablation} presents 
comprehensive ablation studies analyzing frontier node sampling strategies 
and inertial gradient tree expansion schemes. 
Sec.~\ref{suppl:exp} extends experimental validation with path quality analysis, 
real-world Fetch deployment, planning with generated goal images, tree structure visualization, 
and visual ambiguity analysis.
Sec.~\ref{suppl:implement} details implementation including dataset construction, and hyperparameter specifications.

\vspace{-1mm}
\begin{tcolorbox}[
    colback=gray!5,
    colframe=gray!5,
    boxrule=0.05pt,
    sharp corners,
    left=1pt, right=1pt, top=1pt, bottom=1pt,
]
\noindent\textbf{Video Results:} The video results are available at \small{\url{https://sgvr.kaist.ac.kr/Visual-RRT}}.
\end{tcolorbox}

\section{Background of RRT}\label{suppl:rrt}

Rapidly-exploring Random Trees (RRT) is a sampling-based motion planning algorithm that incrementally constructs a search tree to find motion paths in the robot's C-space. The algorithm operates through the following iterative process.

\begin{enumerate}
    \item  \textit{Random Sampling}: Sample a random configuration $q_\text{rand}$ from the C-space.
    \item \textit{Nearest Node Selection}: Find the nearest node $q_\text{near}$ in the current tree to $q_\text{rand}$.
    \item \textit{Tree Extension}: Extend from $q_\text{near}$ toward $q_\text{rand}$ by a fixed step size $\epsilon$ to generate a new configuration $q_\text{new}$.
    \item \textit{Feasibility Check}: If $q_\text{new}$ is feasible (\textit{e.g.}, collision or joint-limit violations), add it to the tree as a child of $q_\text{near}$.
    \item \textit{Termination}: Repeat until a node reaches the goal region.
\end{enumerate}

\vspace{2mm}
\noindent \textbf{Goal-biased Exploitation.}
To balance exploration and exploitation, RRT typically employs goal biasing: with a certain probability, $q_\text{rand}$ is set to the goal configuration $q_\text{goal}$ instead of being sampled uniformly. This biasing steers tree growth toward the goal region, significantly accelerating convergence. However, this strategy fundamentally relies on the availability of an explicit goal configuration $q_\text{goal}$, typically specified as numerical joint angles. 
In visual-goal planning scenarios where goals are provided as images rather than configurations, this dependency poses a key challenge, as there is no explicit $q_\text{goal}$ to bias toward.
\section{Extended Ablation Studies}\label{suppl:ablation}
\subsection{Frontier-based Exploration-Exploitation}
\noindent \textbf{Truncated Geometric Distribution.} In \cref{tab:suppl_geom_distribution}, we analyze the effectiveness of different frontier node selection strategies on visual planning performance. In our method, the truncated geometric distribution parameter $\kappa \in [0, 1)$ controls the frontier selection bias: smaller $\kappa$ focuses on sampling visually promising nodes. Rows (a-c) show that average success rates improve with higher $\kappa$. In particular, at $2.5$ rad, $\kappa=0.9$ maintains higher success rates compared with $\kappa=0.5$. These results indicate that stochastic diversity in frontier selection helps navigate distant-goal scenarios where gradient-based exploitation is prone to local minima.

\vspace{1mm}
\noindent \textbf{Alternative Frontier Selection.} As shown in \cref{tab:suppl_geom_distribution}, we evaluate uniform sampling (d) and Top-K selection (e) as alternative frontier node selection methods for comparison. Uniform sampling selects nodes without 
considering visual loss, while Top-K deterministically chooses the K best nodes.
The uniform sampling strategy demonstrates limited effectiveness, as it allocates computational effort across all nodes without prioritizing visually promising regions.
Top-K selection shows reasonable planning results comparable to $\kappa\!=\!0.5$. This similarity reflects their shared behavior where both heavily concentrate selection on visually promising nodes.
Accordingly, they exhibit similar challenges at larger distances ($2.0$ and $2.5$ rad), 
compared to $\kappa\!=\!0.9$.

\begin{table}[]
\renewcommand{\tabcolsep}{0.8mm}
\renewcommand{\arraystretch}{1.0}
\centering
\footnotesize
\caption{
\textbf{Ablation on frontier sampling strategies.} Success rates (\%) across 
C-space distance bins for frontier node selection methods. 
Higher $\kappa$ achieves better performance through stochastic diversity. Uniform sampling (d) 
and Top-K (e) show limitations compared to $\kappa\!=\!0.9$ at larger distances (2.0 and 2.5 rad).
}
\vspace{-3mm}
\label{tab:suppl_geom_distribution}
\begin{tabular}{cl|ccccc|c}
\toprule
& \multicolumn{1}{l|}{\multirow{2}{*}{Frontier Sampling}} &
\multicolumn{5}{c|}{Target distance bins (rad)} &
\multicolumn{1}{c}{\multirow{2}{*}{Avg.}} \\
&  & $0.5$ & $1.0$  & $1.5$  & $2.0$ & $2.5$ & \\
\midrule
(a)  & Trunc. geometric ($\kappa$=0.9) & \textbf{87.3} & 87.5 & \textbf{85.2} & \textbf{76.2} & \textbf{62.7} & \textbf{79.8} \\
(b)  & Trunc. geometric ($\kappa$=0.7) & 86.7 &   \textbf{88.3} &   83.4 &   74.0 &   55.0 &   77.5 \\
(c)  & Trunc. geometric ($\kappa$=0.5) & 86.3 &   85.7 &   82.6 &   72.5 &   54.2 &   76.2 \\
\cmidrule{1-8}
(d)  & Uniform    &   17.3 &    3.5 &    1.3 &    0.3 &    0.8 &    4.7 \\
(e)  & Top-K &   86.8 &   86.7 &   82.7 &   71.3 &   53.2 &   76.1 \\
\bottomrule
\end{tabular}
\vspace{-1mm}
\end{table}

\subsection{Inertial Gradient Tree Expansion}
\begin{table}[t]
\renewcommand{\tabcolsep}{0.8mm}
\renewcommand{\arraystretch}{1.0}
\centering
\footnotesize
\caption{\textbf{Ablation on optimizer state inheritance.}
Success rates (\%) across C-space distances for different inherited optimizer states in vRRT.
Naive gradient descent without any optimizer state (a) is less effective at exploiting visual gradients.
Inheriting a simple momentum term (b)--(g) substantially improves performance over stateless tree expansion.
Adaptive optimizers that accumulate gradient statistics (h)--(k) further outperform the best momentum-only baseline, with Adam achieving the highest average success rate.}
\vspace{-3mm}
\label{tab:suppl_optim}
\begin{tabular}{cl|ccccc|c}
\toprule
& \multicolumn{1}{l|}{\multirow{2}{*}{Inherited Optimizer State}} &
\multicolumn{5}{c|}{Target distance bins (rad)} &
\multicolumn{1}{c}{\multirow{2}{*}{Avg.}} \\
&  & $0.5$ & $1.0$  & $1.5$  & $2.0$ & $2.5$ & \\
\midrule
(a)  & Naive GD (Eq. 1) &   21.8 &    3.8 &    2.2 &    4.7 &    3.5 &    7.2 \\
(b)  & Momentum ($\mu=0.5$) &   37.8 &   16.3 &   10.1 &   12.8 &   10.8 &   17.6 \\
(c)  & Momentum ($\mu=0.7$) &   54.0 &   28.6 &   21.3 &   22.2 &   16.7 &   28.5 \\
(d)  & Momentum ($\mu=0.8$) &   68.8 &   38.6 &   32.9 &   26.7 &   15.3 &   36.5 \\
(e)  & Momentum ($\mu=0.85$) &   73.2 &   49.4 &   38.7 &   31.5 &   22.0 &   43.0 \\
(f)  & Momentum ($\mu=0.9$) &   73.8 &   60.4 &   46.3 &   37.7 &   23.5 &   48.3 \\
(g)  & Momentum ($\mu=0.95$) &   72.3 &   67.5 &   57.1 &   43.7 &   32.2 &   54.6 \\
\cmidrule{1-8}
(h)  & Adaptive gradient~\cite{duchi2011adaptive}  &   83.3 & 80.4 &   71.9 &   59.2 &   43.5 &   67.7 \\
(i)  & RMSProp~\cite{tieleman2012lecture} &   89.5 &   80.2 &   70.6 &   53.8 &   42.7 &   67.4 \\
(j)  & Lion~\cite{chen2023symbolic}  &  87.8 &   86.3 &   81.6 &   64.3 &   47.3 &   73.5 \\
(k)  & Adam~\cite{Kingma2015AdamAM} (Eq. 5) & 87.3 & 87.5 & 85.2 & 76.2 & 62.7 & 79.8 \\
\bottomrule
\end{tabular}
\vspace{-3mm}
\end{table}

In the main paper, we instantiate inertial gradient tree expansion with Adam optimizer states (\textit{i.e.}, first- and second-moment estimates with iteration steps).
The first-moment estimate in Adam plays the role of momentum along each branch of the tree, which motivates the term ``inertial'' in our formulation.
In this supplementary material, we further compare several alternative optimization states within the same state-inheritance framework.

\vspace{1mm}
\noindent \textbf{Momentum in Tree Expansion.}
We examine the effectiveness of inertia using a simpler momentum formulation.
In this variant, each node stores a velocity vector $u_p$ that is inherited from parent to child and updated with Polyak momentum~\cite{polyak1964some}. Given a parent configuration $q_p$ and its velocity $u_p$, the child inherits velocity $u_{\text{new}}$ and configuration $q_{\text{new}}$ as:
\vspace{-5mm}
\begin{align}
u_{\text{new}} &= \mu\, u_p + (1 - \mu)\, \nabla_q \mathcal{L}_\text{render}(q_p), \\
q_{\text{new}} &= q_p - \alpha\, u_{\text{new}},
\label{suppl:eq:sgdm}
\end{align}
where $\mu \in [0,1)$ is the momentum coefficient and $\alpha$ is the step size. 
In \cref{tab:suppl_optim}, row (a) shows that naive gradient descent (Naive GD) fails to 
leverage visual gradients effectively without optimization inertia. 
Meanwhile, rows (b)--(g) show that inheriting this simple momentum term along the tree improves visual motion planning performance over stateless gradient descent, demonstrating the importance of propagating optimizer states across tree expansions in our framework.

\vspace{1mm}
\noindent \textbf{Alternative Optimization State Inheritance.}
Our framework only requires each node to maintain a compact summary of its gradient history during tree expansion, which can be realized through various optimizer designs.
Thus, we further evaluate several history-aware optimization variants (Adaptive gradient~\cite{duchi2011adaptive}, RMSProp~\cite{tieleman2012lecture}, Lion~\cite{chen2023symbolic}, Adam~\cite{Kingma2015AdamAM}) by plugging them into the same state-inheritance mechanism.
In particular, Adaptive gradient and RMSProp accumulate per-parameter squared gradients to adapt the effective step size, while Lion and Adam additionally maintain directional statistics over past gradients.
As shown in \cref{tab:suppl_optim} (h)--(k), these history-aware instantiations substantially outperform naive gradient descent (Naive GD), indicating that propagating optimizer states along the tree is more important than the exact choice of optimizer.
Among these variants, Adam achieves the highest average success rate, supporting our default design choice in the main paper.

\section{Extended Motion Planning Evaluation}\label{suppl:exp}

We provide comprehensive analysis of vRRT's visual-goal motion planning performance.
We first validate path quality and real-world deployment, then analyze our tree structure and computational efficiency.
Finally, we discuss inherent limitations arising from visual ambiguity.

\subsection{Path Quality Analysis}
\begin{table}[t]
\renewcommand{\tabcolsep}{0.8mm}
\renewcommand{\arraystretch}{1.1}
\centering
\footnotesize
\caption{\textbf{Comparing RRT$^\ast$ path lengths across recovered goals.}
RRT$^\ast$ paths to vRRT-recovered goals are consistently longer, suggesting the methods solve different problem subsets: vRRT addresses more challenging instances requiring longer paths, while Dr.Robot~\cite{drrobot} succeeds on 
shorter-distance cases.}
\vspace{-3mm}
\label{tab:suppl_mp_pathlength}
\begin{tabular}{ll|ccccc|c}
\toprule
\multicolumn{1}{l}{\multirow{2}{*}{Robot}} & \multicolumn{1}{l|}{\multirow{2}{*}{RRT$^\ast$}} &
\multicolumn{5}{c|}{Target distance bins (rad)} &
\multicolumn{1}{c}{\multirow{2}{*}{Avg.}} \\
&  & $0.5$ & $1.0$  & $1.5$  & $2.0$ & $2.5$ & \\
\midrule
\multirow{2}{*}{Franka}  
    & \makecell[l]{w/ Dr.Robot goal} & 0.50 & 0.99 & 1.40 & 1.90 & 2.49 & 1.46 \\
    & \makecell[l]{w/ vRRT goal}     & 0.50 & 1.00 & 1.55 & 2.08 & 2.63 & \textbf{1.55} \\
\midrule
\multirow{2}{*}{UR5e}  
    & \makecell[l]{w/ Dr.Robot goal} & 0.50 & 1.02 & 1.52 & 2.00 & 2.55 & 1.52 \\
    & \makecell[l]{w/ vRRT goal}     & 0.50 & 1.01 & 1.55 & 2.10 & 2.59 & \textbf{1.55} \\
\midrule
\multirow{2}{*}{Fetch}  
    & \makecell[l]{w/ Dr.Robot goal} & 0.47 & 0.98 & 1.49 & 1.93 & 2.46 & 1.47 \\
    & \makecell[l]{w/ vRRT goal}     & 0.49 & 1.00 & 1.52 & 2.07 & 2.57 & \textbf{1.53} \\
\bottomrule
\end{tabular}
\vspace{-2mm}
\end{table}

\begin{figure*}[]
  \centering
    \includegraphics[width=1\linewidth,trim={0cm 2cm 0cm 0cm},clip]{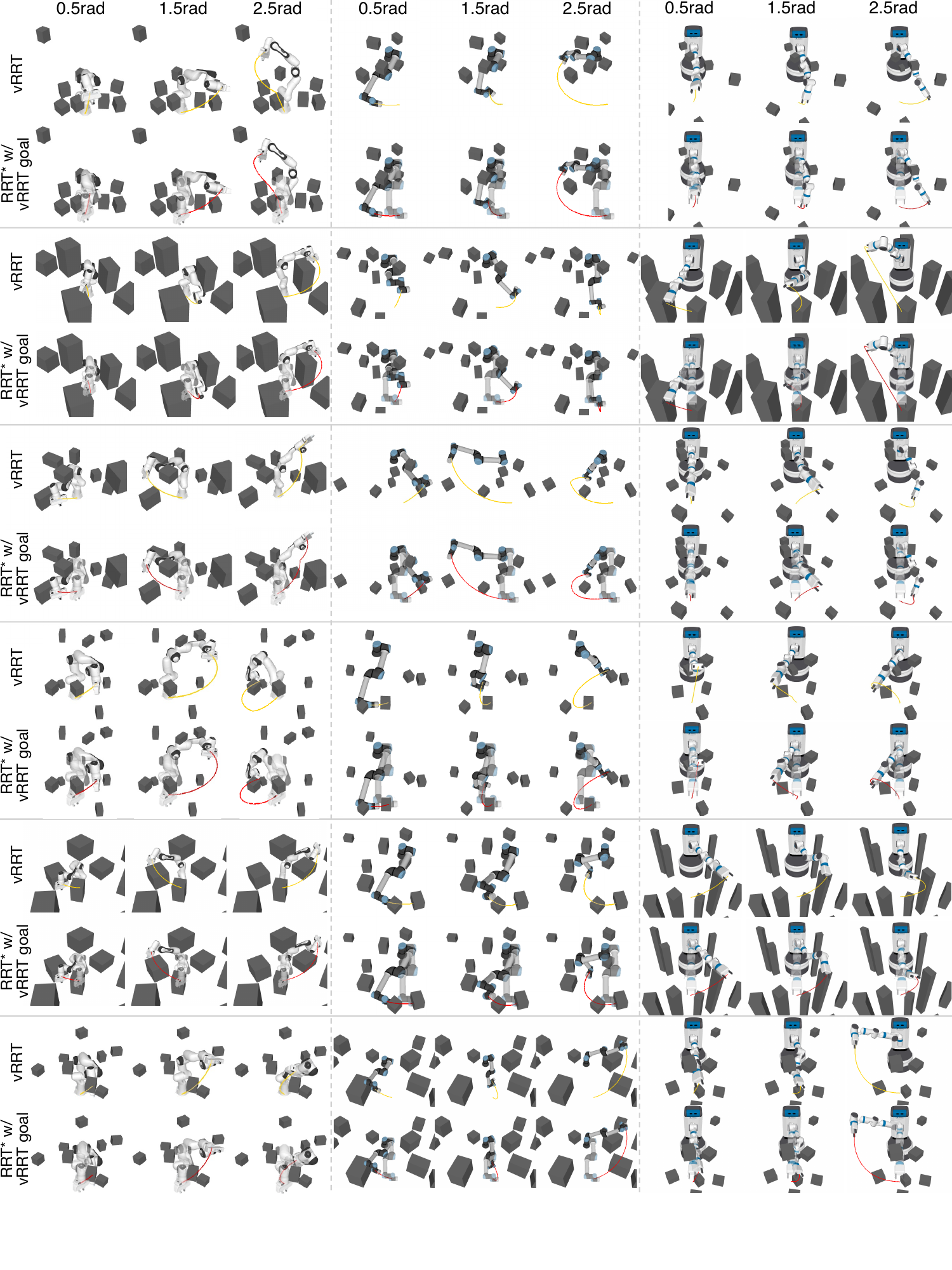}
    \vspace{-7mm}
\caption{\textbf{Path comparison between vRRT and RRT$^\ast$.} 
Paths from vRRT (yellow) and RRT$^\ast$ (red) across three robots in six 
scenes each. vRRT produces paths similar to RRT$^\ast$ despite using only visual objectives. Videos are provided in the project page.}
    \label{fig:suppl_path_comparison}
\vspace{-4mm}
\end{figure*}

In Tab.~1 of the main paper, vRRT achieves higher success rates than the baselines, but yields moderately longer paths than Dr.Robot + RRT$^\ast$. To determine whether this difference reflects planning inefficiency or differences in problem coverage, we analyze the recovered goal configurations of both methods by running RRT$^\ast$ from the same start configuration to each recovered goal. Tab.~\ref{tab:suppl_mp_pathlength} shows that goals recovered by vRRT consistently induce longer RRT$^\ast$ paths than those recovered by Dr.Robot. This suggests that the two methods succeed on different subsets of problems: vRRT solves more challenging instances that inherently require longer paths, whereas Dr.Robot tends to succeed on easier ones. Therefore, the longer paths reported for vRRT in Tab.~1 are better explained by broader problem coverage than by planning inefficiency.
Fig.~\ref{fig:suppl_path_comparison} presents representative trajectories from vRRT and RRT$^\ast$ across 18 scenes. Despite relying only on visual objectives without explicit goal configurations, vRRT consistently produces paths that closely match the geometric structure of the RRT$^\ast$ reference solutions. This further supports that our frontier-based exploration--exploitation strategy can recover C-space-efficient paths from visual goals.

\subsection{Real-world Validation}
\begin{figure*}[]
  \centering
    \includegraphics[width=1\linewidth,trim={0cm 18.5cm 0cm 0cm},clip]{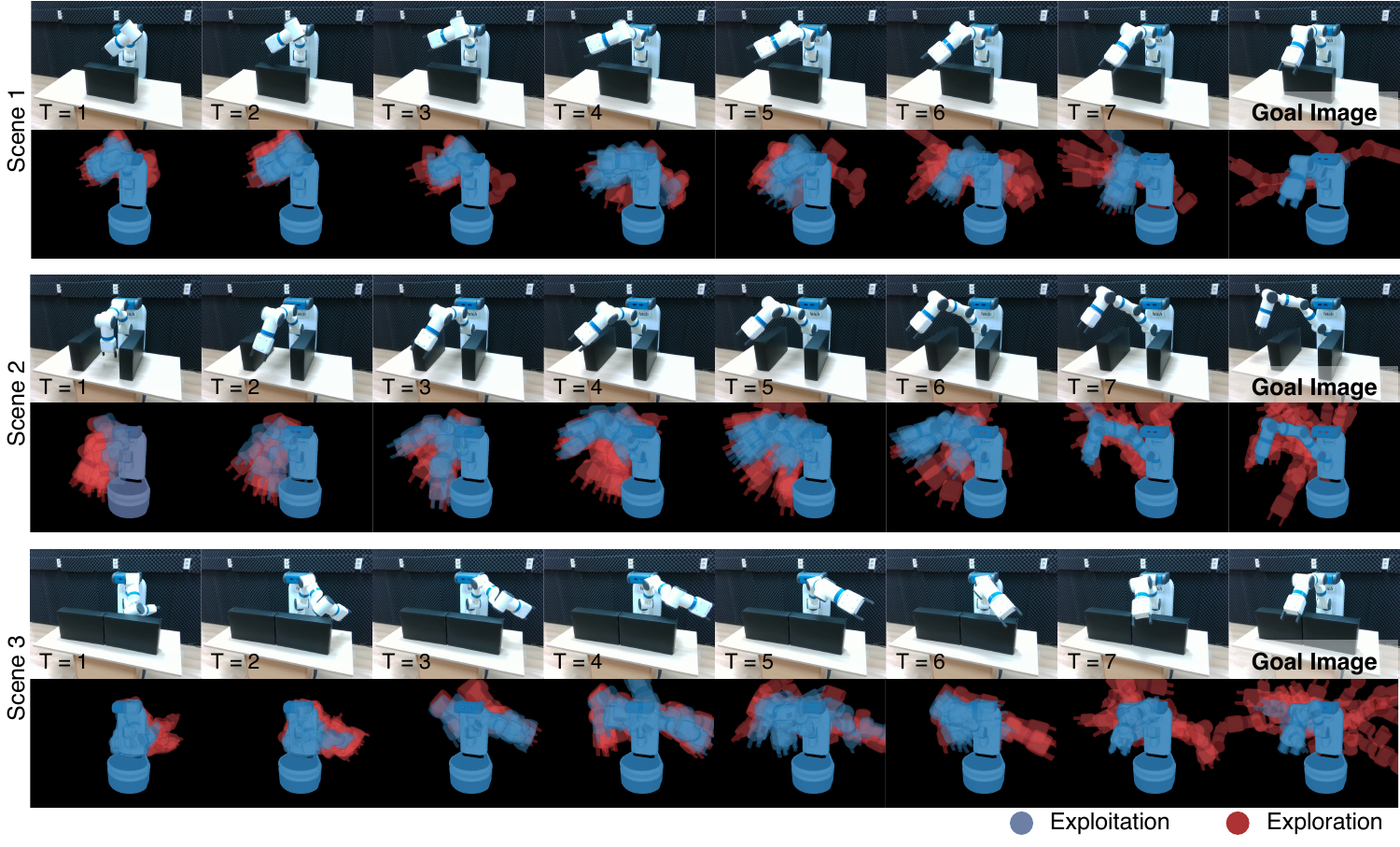}
        \vspace{-8mm}
\caption{\textbf{Fetch deployment across three scenes.} 
Representative executions from three different obstacle 
configurations. Top row shows executed paths reaching visual goals. 
Bottom row shows planning dynamics where exploitation samples (blue) 
follow visual gradients while exploration samples (red) maintain coverage. The videos are provided in the project page.}
\label{fig:suppl_realfetch}
        \vspace{-5mm}  
\end{figure*}

\begin{figure}[]
\vspace{1mm}
  \centering
    \includegraphics[width=1\linewidth,trim={0cm 17cm 0cm 0cm},clip]{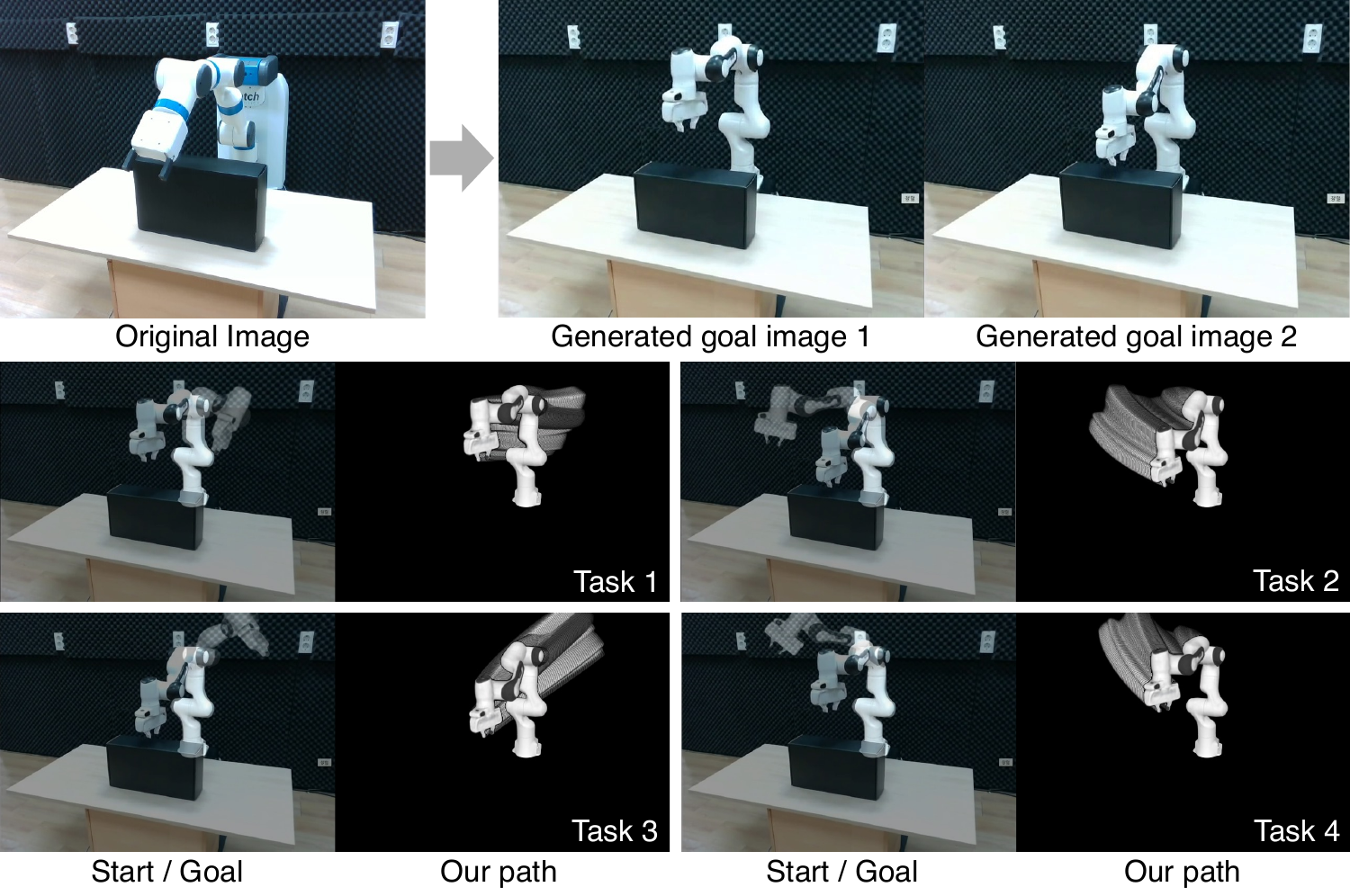}
        \vspace{-7mm}
\caption{\textbf{Visual-goal motion planning with generated goal images.} 
To demonstrate broader applicability, we generate goals by prompting an 
generation model~\cite{google_gemini} to inpaint a Franka robot into Fetch scenes 
while maintaining background. Four tasks show paths matching synthesized poses, 
demonstrating potential to bridge generative models with executable planning.}\label{fig:suppl_gemini}
        \vspace{-5mm}
\end{figure}

\begin{table}[t]
\renewcommand{\tabcolsep}{3.8mm}
\renewcommand{\arraystretch}{0.9}
\centering
\footnotesize
\caption{\textbf{Real-world Fetch experiments.} 
Success counts across three scenes with 25 tasks each. 
vRRT achieves 80\% success rate (60/75) in real-world deployment, 
demonstrating potential for sim-to-real transfer in visual-goal motion planning.}
\vspace{-3mm}
\label{tab:suppl_realfetch}
\begin{tabular}{l|ccc|c}
\toprule
& Scene 1 & Scene 2  & Scene 3  & Total\\
\midrule
Tasks     & 25 & 25 & 25 &  75 \\
Successes & 18 & 24 & 18 &  60 \\
\bottomrule
\end{tabular}
\vspace{-4mm}
\end{table}

We deploy vRRT on a physical Fetch robot across three scenes with varying obstacle configurations: Scene 1 with one obstacle and Scenes 2-3 with two obstacles each. 
For each scene, we design 25 tasks requiring the robot to 
reach goals on either side or front/back of obstacles, verified for reachability via RRT. 
Tab.~\ref{tab:suppl_realfetch} shows vRRT achieves 80\% success rate 
(60/75 tasks), demonstrating promising sim-to-real transfer. 
Fig.~\ref{fig:suppl_realfetch} shows representative executions and planning, successfully guiding the robot to visual goals despite domain gap. 
Failures occur when RGB appearance similarities between robot 
and scene elements create misleading visual gradients.

\vspace{-0.5mm}
\subsection{vRRT with Generated Goal Images}
To further demonstrate practical generality of our visual-goal planning framework, we introduce an additional visual-goal motion planning experiment where the goal image is generated by an image generation model from a natural-language description. 
While recent advances enable synthesis of visually realistic robot scenes, 
generating motion paths that align with these synthesized frames remains challenging~\cite{lee2025dynscene}.
We use an image generation model~\cite{google_gemini} to replace the Fetch 
robot in Fig.~\ref{fig:suppl_gemini} with a Franka Emika Panda while preserving background and viewpoint. 
To guide robot appearance, we provide our differentially rendered Franka as a conditioning image. 
For each synthesized goal, we randomly sample start configurations and plan paths using vRRT. 
Fig.~\ref{fig:suppl_gemini} shows successful recovery of target configurations, 
demonstrating vRRT's potential to bridge generative models with executable planning.

\subsection{Tree Structure Analysis}
\begin{figure*}[]
  \centering
    \includegraphics[width=1\linewidth,trim={0cm 24.5cm 0cm 0.0cm},clip]{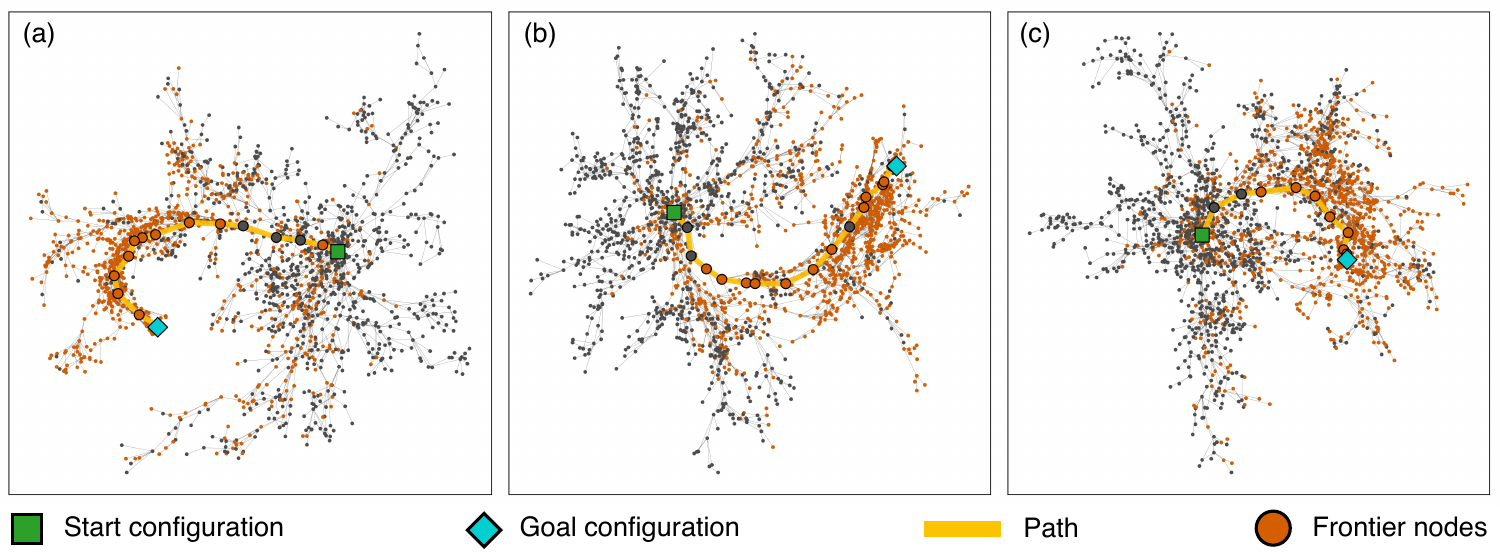}
        \vspace{-7mm}
\caption{\textbf{Tree structure with frontier node visualization.}
Configuration space projected onto 2D via PCA for three planning instances (a-c). 
Frontier nodes (orange) represent configurations with low visual loss selected for effective expansion, while gray nodes indicate non-frontier configurations. 
The spatial distribution shows frontier nodes progressively concentrate toward 
the goal region, validating that our frontier-based sampling effectively prioritizes visually promising areas for exploration and exploitation.}
\label{fig:suppl_tree_frontier}
        \vspace{-1mm}
\end{figure*}

\begin{figure*}[]
  \centering
    \includegraphics[width=1\linewidth,trim={0cm 24.5cm 0cm 0cm},clip]{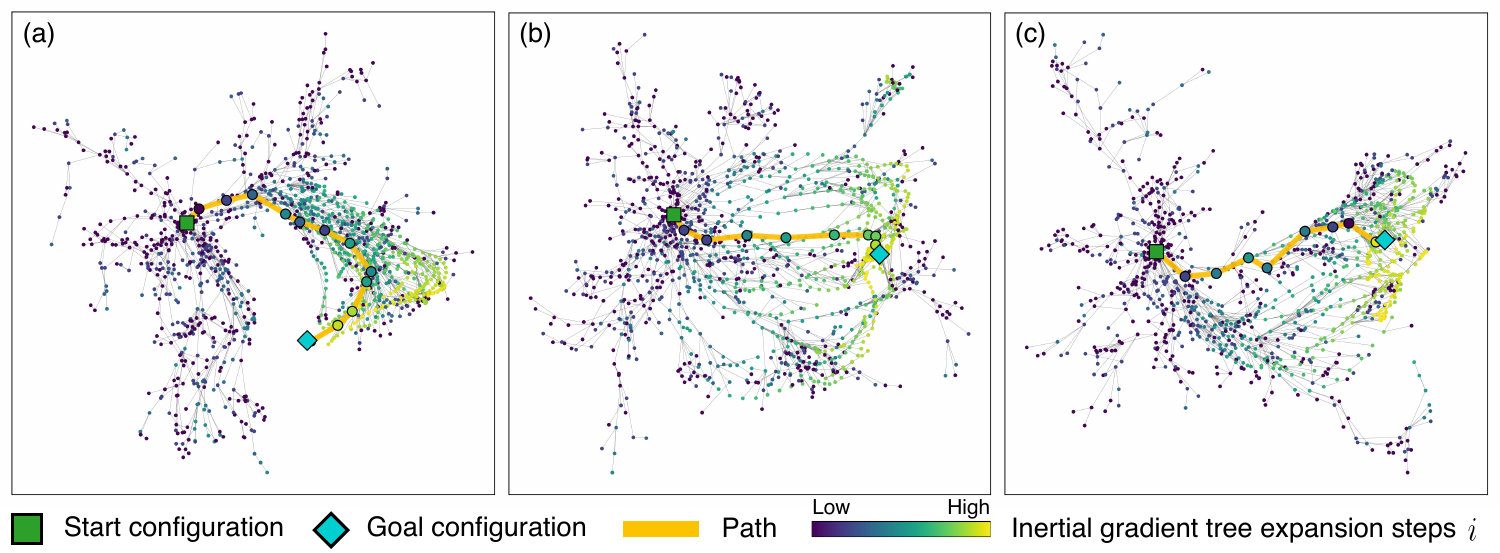}
        \vspace{-7mm}
\caption{\textbf{Tree structure colored by inertial gradient expansion steps.}
Robot joint configurations visualized in 2D via PCA across three planning instances (a-c).
Node colors indicate inertial gradient tree expansion steps $i$: 
low values (purple) represent nodes primarily reached through exploration, while high values (yellow) indicate nodes refined through extensive gradient optimization.
Progressive color transitions show that state inheritance enables continuous optimization across tree branches. Notably, convergence patterns vary across instances: (a) exhibits a cohesive gradient flow where optimization steps increase smoothly toward the goal, whereas (b) and (c) show multiple distinct optimization branches reaching the target.}
\label{fig:suppl_tree_inertial}
\vspace{-4mm}
\end{figure*}

To visualize how vRRT explores the configuration space during planning, we project robot joint configurations of our tree structure to 2D subspace via Principal Component Analysis (PCA)~\cite{mackiewicz1993principal}.
Each node in the visualization represents a single configuration from the robot C-space.

\vspace{1mm}
\noindent\textbf{Frontier selection.}
In Fig.~\ref{fig:suppl_tree_frontier}, we visualize the effectiveness of frontier-based exploration-exploitation scheme during tree expansion. Frontier nodes (orange), sampled via truncated geometric 
distribution based on visual loss ranking, concentrate toward the promising goal region, 
while non-frontier nodes (gray) maintain exploration coverage. 
This spatial distribution validates that our rank-based probabilistic sampling strategy effectively biases 
tree growth toward visually aligned configurations.

\vspace{1mm}
\noindent\textbf{Inertial gradient expansion.}
Fig.~\ref{fig:suppl_tree_inertial} shows trees colored by inertial gradient expansion steps $i$. 
Low values (purple) indicate configurations primarily 
discovered through random steering exploration, while high values (yellow) represent configurations generated through our inertial gradient tree expansion. 
The tree structure reveals two key characteristics. First, exploration establishes broad coverage from the start, while gradient-steered nodes concentrate near the goal. 
Second, the convergence patterns differ across instances: 
(a) exhibits a cohesive stream of nodes flowing toward the goal, 
whereas (b, c) show multiple distinct branches extending to the target. 
These different convergence patterns suggest that inertial expansion can flexibly support both steady refinement along a dominant branch and parallel pursuit of multiple promising routes.
These observations validate inertial gradient tree expansion: 
smooth color transitions along branches confirm continuous optimization trajectories through state inheritance, 
while the coexistence of low and high $i$ nodes demonstrates effective balance between exploration and exploitation throughout planning. 

\subsection{Computational Cost}
\begin{table}[t]
\renewcommand{\tabcolsep}{2.1mm}
\renewcommand{\arraystretch}{0.9}
\centering
\footnotesize
\caption{\textbf{Computational cost per iteration.} 
Average time allocation (\%) when expanding 32 nodes per iteration across 
three robot platforms. Rendering dominates the cost indicating that reducing Gaussians directly improves vRRT efficiency.}
\vspace{-3mm}
\label{tab:suppl_runtime}
\begin{tabular}{lcccc}
\toprule
\makecell{Robot \\(\# of Gaussians)} & Rendering & \makecell{Frontier\\sampling} & \makecell{RRT\\operations} & \makecell{Collision\\checking} \\
\midrule
Franka (52,393) & 80.7 & 2.6 & 11.9 & 4.7  \\
UR5e (16,327) & 60.6 & 4.0 & 21.6 & 13.8  \\
Fetch (70,335) & 84.7 & 1.9 & 9.4 & 4.1  \\
\bottomrule
\end{tabular}
\vspace{-3mm}
\end{table} 
We analyze vRRT's computational cost during tree expansion, where 32 nodes 
are inserted per iteration through exploration and exploitation. 
Tab.~\ref{tab:suppl_runtime} shows the average time breakdown across components.
Differentiable rendering dominates (60.6-84.7\%), as each gradient-based 
expansion requires evaluating visual loss and backpropagating through the 
Gaussian Splatting model. This cost scales with the number of Gaussians 
representing the robot. Consequently, robots with more Gaussians exhibit 
higher planning times in Tab.1 of the main paper, indicating that reducing Gaussian count 
directly improves efficiency.
Our algorithmic components, frontier-based sampling and inertial gradient 
tree expansion, account for moderate computational cost. Frontier sampling 
requires 1.9-4.0\% through sorting cached rendering losses. RRT operations including random sampling, nearest-neighbor queries, steering, and tree rewiring, scale with tree size but remain independent of Gaussian 
resolution. Further optimization may be possible through improved data structures 
for nearest-neighbor queries.
Collision checking is performed via MuJoCo and could be replaced 
with alternative methods depending on application requirements.

\subsection{Visual Ambiguity in Goal Specification}
\begin{figure}[]
  \centering
    \includegraphics[width=1\linewidth,trim={0cm 20cm 0cm 0cm},clip]{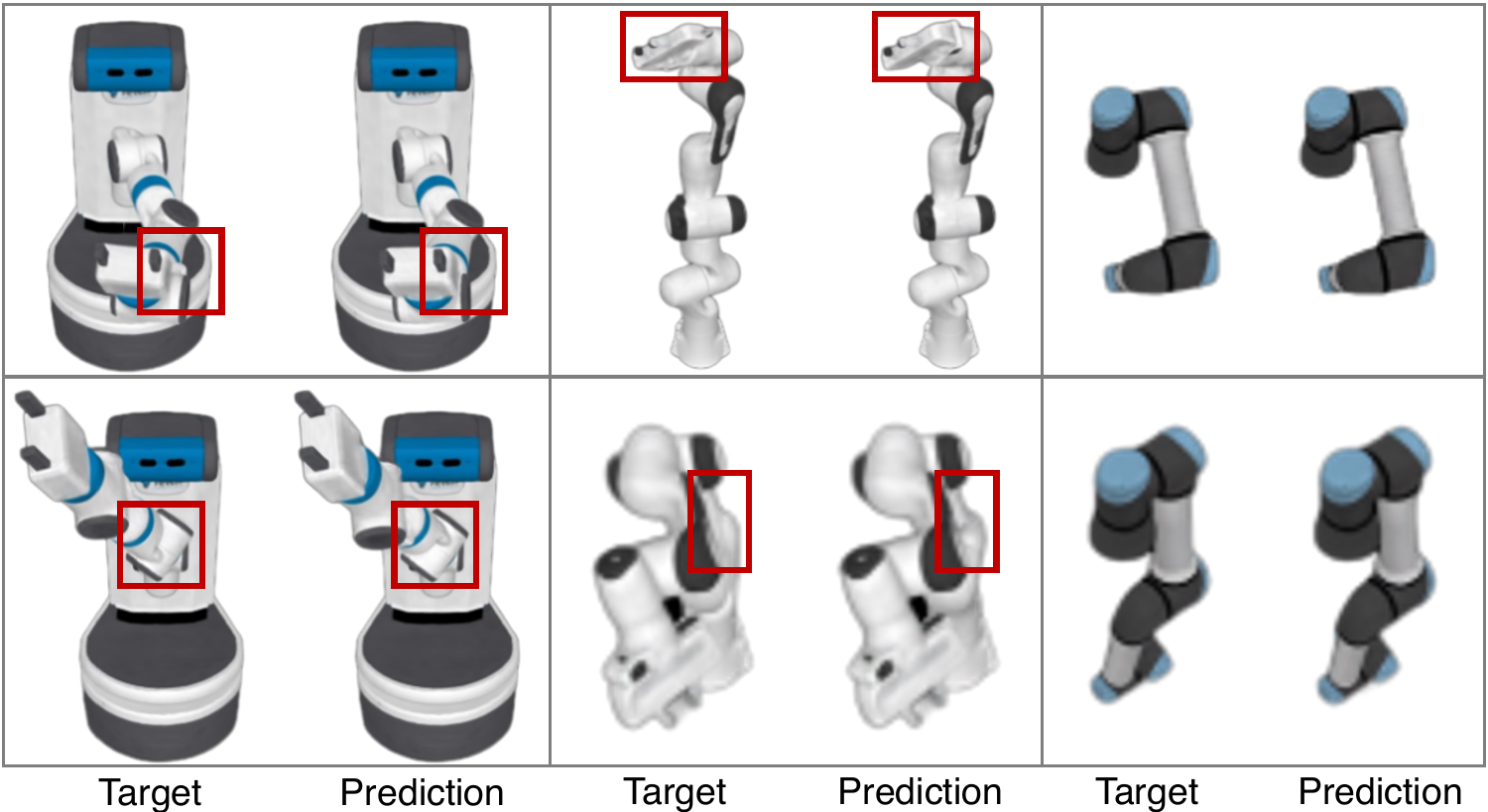}
        \vspace{-6mm}
\caption{\textbf{Representative failure cases.} Target pose images (left) visually resemble predictions (right) but differ in joint space. 
Fetch and Franka: symmetric poses produce identical silhouettes. UR5e: self-occlusion allows hidden joints to rotate without visual change. These illustrate the inherent 
difficulty of visual-goal planning.}
\label{fig:suppl_limitation}
        \vspace{-4mm}
\end{figure}

By integrating sampling-based exploration with visual-gradient exploitation, vRRT inherits RRT's robust exploration that systematically covers the search space given sufficient iterations, while extending to visual-goal specifications. 
Our method demonstrates effective performance in both simulation and real-world settings by leveraging exploration to escape local minima that trap gradient-based methods. 
However, visual ambiguity presents a limitation where 
exploration cannot disambiguate visually indistinguishable configurations.
Fig.~\ref{fig:suppl_limitation} shows representative failure cases where vRRT converges to visually matching but configuration-space incorrect poses. 
Fetch and Franka demonstrate symmetric poses producing identical silhouettes; UR5e shows severe self-occlusion making distinct configurations appear identical. 
When multiple configurations are visually indistinguishable, determining the true goal from observation alone becomes infeasible. 
This represents a fundamental limitation of single-view specification, addressable through multi-view fusion or perceptual features rendering.

\section{Additional Implementation Details}\label{suppl:implement}

\subsection{Dataset Construction}
We construct two synthetic test datasets for motion planning and pose 
reconstruction using the same sampling process, differing only in scene construction. 
For each robot platform, we define a fixed 
canonical pose as the start configuration $q_{\text{start}}$ for all test 
queries, following the same convention as Dr.Robot~\cite{drrobot}, shown 
as faded poses in Fig.~\ref{fig:suppl_path_comparison}.

\vspace{1mm}
\noindent \textbf{Goal configuration sampling.} 
We generate goal configurations by randomly perturbing all revolute joints 
from the canonical pose within their physical limits. 
We verify feasibility of each sampled goal by running RRT~\cite{rrt} 
from the canonical pose with a 30-second time budget. Queries are discarded if 
RRT fails to find a collision-free path or if executing the trajectory 
in MuJoCo reveals collisions. Validated queries are stratified by configuration-space 
distance $\|q_{\text{start}} - q_{\text{goal}}\|_2$ into bins at $[0.5, 1.0, 1.5, 2.0, 2.5]$ 
radians. 

\vspace{1mm}
\noindent \textbf{Scene construction.}
We construct six distinct scenes per robot: obstacle environments 
for motion planning and obstacle-free environments for pose reconstruction. 
For motion planning, each environment contains ten box-shaped obstacles 
with random dimensions drawn from predefined ranges ($3$-$20$ cm for Franka 
and UR5e; $5$-$70$ cm for Fetch). We employ a two-stage placement strategy: 
(1) obstacles are positioned near robot links in the canonical pose without 
intersecting geometry; (2) remaining obstacles are placed randomly within 
the workspace. For pose reconstruction, scenes remain obstacle-free to 
isolate visual goal matching from collision avoidance.

\vspace{1mm}
\noindent \textbf{Dataset statistics.} 
Our dataset comprises 100 validated queries per distance bin per robot. 
For motion planning, this yields 500 queries per robot across five bins 
with six scene variations each, totaling 9,000 problems. For pose reconstruction, 
we generate 500 queries per robot in obstacle-free scenes, totaling 1,500 problems.

\subsection{Pipeline Details}
\noindent \textbf{Robot Gaussian Construction.} We construct differentiable robot representations for all three platforms 
following Dr.Robot~\cite{drrobot}. Starting from MJCF XML descriptions in 
MuJoCo~\cite{todorov2012mujoco}, which define kinematic structure, joint 
limits, and collision geometries, we sample diverse joint configurations 
within physical limits and render multi-view RGB images with varying camera 
poses. Franka and UR5e use standard MuJoCo models, while we custom a MJCF for Fetch following MuJoCo conventions.

\vspace{1mm}
\noindent\textbf{Differentiable robot rendering.}
We train 3D Gaussian Splatting models~\cite{kerbl2023} to represent each robot. 
The optimization follows standard Gaussian Splatting training, learning positional 
and rotational parameters per primitive. The resulting models combine forward 
kinematics with implicit linear blend skinning~\cite{loper2023}, enabling 
differentiable rendering at arbitrary joint configurations.

\vspace{1mm}
\noindent \textbf{Rendering during Planning.} 
We render the robot at the same resolution as the goal image (480$\times$480) and compute a per-pixel $L_2$ loss between the rendered and goal images.
The pre-trained Gaussian model provides visual gradients to the configuration space without additional training during planning. Since the visual gradient $\nabla_q \mathcal{L}_{\text{render}}(q_p)$ at a parent node $q_p$ remains constant throughout a planning iteration, we implement a caching mechanism to avoid redundant rendering operations during tree expansion.

\vspace{1mm}
\noindent \textbf{Optimization State and Settings.} 
The root node (starting configuration) and nodes generated via random steering (exploration) are initialized with zero optimization states: $m \!=\! \mathbf{0}$, $v \!=\! \mathbf{0}$, and $i \!=\! 0$. In contrast, nodes created through visual-gradient steering (exploitation) inherit and update the optimization states from their parent nodes as described in Eq.~(3)-(5) of the main paper, enabling momentum-consistent gradient descent trajectories across tree branches.
We use $M\!=\!200$, $r\!=\!0.3$, $\eta\!=\!0.7$, $\beta_1\!=\!0.9$, and $\beta_2\!=\!0.9$ for experiments. For real-world settings, we use RealSense factory intrinsics and estimate camera-to-robot extrinsics via hand-eye calibration, assuming fixed placement.

{
    \small
    \bibliographystyle{ieeenat_fullname}
    \bibliography{main}
}
\end{document}